\documentclass[lettersize,journal]{IEEEtran}
\usepackage{amsmath,amsfonts}
\usepackage{algorithmic}
\usepackage{algorithm}
\usepackage{array}
\usepackage{textcomp}
\usepackage{stfloats}
\usepackage{url}
\usepackage{verbatim}
\usepackage{graphicx}
\usepackage{cite}

\usepackage[algo2e,ruled,linesnumbered]{algorithm2e}
\usepackage{caption}
\usepackage{subcaption}
\usepackage{multirow}
\usepackage{booktabs}
\usepackage{enumitem}
\usepackage{lipsum}

\def\method{DisenSemi}
\usepackage{color}

\begin{document}

\title{DisenSemi: Semi-supervised Graph Classification via Disentangled Representation Learning}


\author{Yifan Wang, Xiao Luo, Chong~Chen,~\IEEEmembership{Member,~IEEE,}~Xian-Sheng~Hua,~\IEEEmembership{Fellow,~IEEE,}\\ Ming Zhang and Wei Ju,~\IEEEmembership{Member,~IEEE}% <-this % stops a space
\IEEEcompsocitemizethanks{
\IEEEcompsocthanksitem Yifan Wang is with School of Information Technology $\&$ Management, University of International Business and Economics, Beijing, 100029, China.
(yifanwang@uibe.edu.cn)
\IEEEcompsocthanksitem Xiao Luo is with Department of Computer Science, University of California, Los Angeles, 90095, USA.
(xiaoluo@cs.ucla.edu)
\IEEEcompsocthanksitem Chong Chen and Xian-Sheng Hua are with Terminus Group, Beijing 100027, China. 
(chenchong.cz@gmail.com, huaxiansheng@gmail.com)
\IEEEcompsocthanksitem Ming Zhang is with School of Computer Science, National Key Laboratory for Multimedia Information Processing, Peking University-Anker Embodied AI Lab, Peking University, Beijing, 100871, China.
(mzhang$\_$cs@pku.edu.cn)
\IEEEcompsocthanksitem Wei Ju is with College of Computer Science, Sichuan University, Chengdu, 610064, China.
(juwei@pku.edu.cn)
% \IEEEcompsocthanksitem .
\IEEEcompsocthanksitem Corresponding author: Wei Ju
}
\thanks{Manuscript received April 19, 2021; revised August 16, 2021.}}

\markboth{Journal of \LaTeX\ Class Files,~Vol.~14, No.~8, August~2021}%
{Shell \MakeLowercase{\textit{et al.}}: A Sample Article Using IEEEtran.cls for IEEE Journals}


\maketitle

\begin{abstract}
% Graph neural networks (GNNs) have emerged as widely adopted methods for graph classification due to their powerful representation ability. However, in real-world scenarios, labeled data may be scarce or limited. Towards this end, we study the problem of semi-supervised graph classification, which consists of a supervised and an unsupervised model learning from supervised and unsupervised data. Instead of recent works that transfer the entire knowledge from the unsupervised model to the supervised model, we argue that a high-performing transfer should only preserve the suitable semantic that aligns well with the supervised task. 

Graph classification is a critical task in numerous multimedia applications, where graphs are employed to represent diverse types of multimedia data, including images, videos, and social networks. Nevertheless, in the real world, labeled graph data is always limited or scarce. To address this issue, we focus on the semi-supervised graph classification task, which involves both supervised and unsupervised models learning from labeled and unlabeled data. In contrast to recent approaches that transfer the entire knowledge from the unsupervised model to the supervised one, we argue that an effective transfer should only retain the relevant semantics that align well with the supervised task. We introduce a novel framework termed \method{} in this paper, which learns disentangled representation for semi-supervised graph classification. Specifically, a disentangled graph encoder is proposed to generate factor-wise graph representations for both supervised and unsupervised models. Then we train two models via supervised objective and mutual information (MI)-based constraints respectively. To ensure the meaningful transfer of knowledge from the unsupervised encoder to the supervised one, we further define an MI-based disentangled consistency regularization between two models and identify the corresponding rationale that aligns well with the current graph classification task. Experiments conducted on various publicly available datasets demonstrate the effectiveness of our \method{}.
% show that \method{} consistently outperforms state-of-the-art approaches.
\end{abstract}


\section{Introduction}
\label{sec:introduction}
\IEEEPARstart{G}{raphs} are among the most versatile data structures in many real-world multimedia applications across various domains, e.g., biology networks~\cite{borgwardt2005protein}, molecule structures~\cite{hao2020asgn}, citation networks~\cite{wang2022disencite}, social networks~\cite{backstrom2011supervised} and recommender system~\cite{wang2020disenhan}, etc. One fundamental task for graph-structured data is graph classification, which endeavors to capture the characteristics of the entire graph and has become an important research hot spot in numerous multimedia tasks, including chemical compound prediction~\cite{hao2020asgn}, text categorization~\cite{linmei2019heterogeneous} and social network analysis~\cite{backstrom2011supervised}.

To tackle the graph classification problem, many machine learning-based methods are proposed. Graph kernel methods leverage structural patterns such as shortest paths~\cite{borgwardt2005shortest}, subtrees~\cite{shervashidze2011weisfeiler}, subgraphs~\cite{kondor2016multiscale} and graphlets~\cite{shervashidze2009efficient} to measure similarity among graphs. These methods then use the similarity matrix with a kernel-based supervised algorithm, like the Support Vector Machine (SVM), to perform classification. Instead of hand-crafted feature extraction in graph kernel methods, more recently, graph neural networks (GNNs) can extract graph structural features in a supervised end-to-end manner. Despite the superior performance, GNNs often require a substantial amount of labeled data, which can be costly and time-consuming in real-world scenarios~\cite{ju2022ghnn}. 

% For example, as demonstrated in Fig.~\ref{fig:1}, the label of a chemical compound is typically produced with a costly Density Functional Theory (DFT) calculation or several complicated experiments.

% \begin{figure}[t]
%     \centering \includegraphics[width=\linewidth]{figure/fig1.pdf}
%     \caption{Illustration of the molecular complicated annotating procedure. It takes around four to five hours to compute the characteristics of the molecule, which fully demonstrates the graph label scarcity in reality.}
%     \label{fig:1}
% \end{figure}

% and active learning methods~\cite{}
In practice, there always exist massive unlabeled graph samples. Although labels of these graphs are unavailable, the inherent knowledge of the graph may help to enhance GNNs' encoding more expressive and discriminative. Semi-supervised classification methods combine both supervised and unsupervised models where the unsupervised model learns from unlabeled data and can serve as a regularizer. Indeed, there are a handful of semi-supervised works proposed for graph classification, which can be categorized into several types, i.e., pseudo-labeling~\cite{ju2022kgnn,luo2022dualgraph,yue2022label}, consistency learning~\cite{you2020graph,sun2020infograph,you2021graph,ju2022tgnn,hao2020asgn,ju2024focus}. Especially for consistency learning, methods like InfoGraph~\cite{sun2020infograph} and GraphSpa~\cite{ju2024focus} jointly learn supervised and unsupervised models from labeled and unlabeled data. Meanwhile, a popular fashion (i.e., GraphCL~\cite{you2020graph} and JOAO~\cite{you2021graph}) has been developed recently, which first pre-train the model in an unsupervised manner using unlabeled data and then fine-tune the model with labeled data.
% TGNN~\cite{ju2022tgnn}

% which can be divided into two categories, i.e., joint learning methods~\cite{sun2020infograph, hao2020asgn, ju2022tgnn, yue2022label, ju2022kgnn, luo2022dualgraph, luo2023towards, jufocus} %ju2022ghnn, 
% and unsupervised learning methods~\cite{hassani2020contrastive, you2020graph, you2021graph, li2021disentangled, li2022let}. On the one hand, methods like InfoGraph~\cite{sun2020infograph}, TGNN~\cite{ju2022tgnn} and GraphSpa~\cite{jufocus} jointly learn supervised and unsupervised models from both labeled and unlabeled data. On the other hand, a popular fashion (i.e., GraphCL~\cite{you2020graph} and JOAO~\cite{you2021graph}) has been developed recently which first pre-trains the unsupervised model with unlabeled data and then fine-tunes the model with labeled data.

However, these previous semi-supervised graph classification methods still face the following limitations. \textit{(1) Neglect the intricate interactions of latent factors.} In the real world, the formation of graphs often results from the intricate interactions of various latent factors. For instance, a molecular graph may include diverse groups of atoms and bonds that represent distinct functional units~\cite{ying2018hierarchical}. Existing works often overlook the interactions of latent factors and the extracted features are holistic, which harms interpretability and leads to sub-optimal performance for the graph classification task. \textit{(2) Mismatch semantics between supervised and unsupervised tasks.} Supervised and unsupervised tasks on labeled and unlabeled graph data may capture different information or occupy separate semantic spaces, corresponding to distinct graph factors. Simply combining two tasks may lead to a ``negative transfer"~\cite{pan2010survey}. Therefore, we argue that a more efficient way is to disentangle the graph representation learned from unsupervised tasks into distinct latent factors and transfer the corresponding factor-wise information well aligned to semantically enhance the supervised task for semi-supervised graph classification.

% expect an approach that is able to explore different latent factors with unlabeled data and transfer the corresponding information semantically to enhance the graph encoder with labeled data for semi-supervised graph classification.

Recently, disentangled representation learning has gained much attention, aiming at learning factorized representations that are capable of uncovering information about the salient (or explanatory) properties of the data. Moreover, disentangled representations have been demonstrated to be more generalizable and resilient to complex variants, i.e., the learned factors are supposed to be independent and less sensitive to the noises from other factors in the limited observed training data~\cite{bengio2013representation}. For graph-structured data, where multiple heterogeneous relations are always mixed and collapsed into one graph, disentangled representation learning tails to decompose underlying relations and is mainly focused on the supervised link prediction tasks (i.e., recommender systems~\cite{wang2020disenhan,wang2022disenctr}), supervised graph classification tasks~\cite{ma2019disentangled,yang2020factorizable} and unsupervised graph generation/identification process~\cite{li2021disentangled,mo2023disentangled}. However, the semi-supervised method of bridging the supervised and unsupervised tasks remains largely unexplored.

% disentangled representation learning brings great benefits to semi-supervised graph classification. It requires the model to decompose underlying relations, yield interpretable features for downstream supervised and unsupervised tasks, and bridge the corresponding feature between tasks.

% \red{In contrast to existing works that exploit entangled influential factors between supervised and unsupervised graph encoders, our proposed approach identifies different underlying latent factors of the graph data and explicitly regularizes the rationale that fits well between supervised and unsupervised learning.}
Toward this end, we propose \textbf{\method{}} in this paper, a novel \textbf{Disen}tangled graph representation learning framework for \textbf{Semi}-supervised graph classification. This framework enhances the supervised prediction task with the unsupervised representation learning tasks in a semantically meaningful way. In contrast to existing works that directly utilize an entangled representation learned from unlabeled data for prediction, our proposed approach emphasizes regularizing the rationale to explicitly exploit transferable factor information between supervised and unsupervised tasks.
% our proposed approach regularizes the rationale that fits well to explicitly exploit transferable factor information between supervised and unsupervised tasks.} 
% exploit transferable entangled representation from unsupervised to supervised graph encoders, our proposed approach identifies different underlying latent factors of the graph data and explicitly regularizes the rationale that fits well between supervised and unsupervised learning.
Specifically, we propose a disentangled graph encoder that characterizes global-level topological semantics. The encoder first decomposes the entire graph into several factor graphs. Then the factor-wise interpretable graph representation is extracted through multiple channels of the message-passing layer. Each channel is specifically designed to propagate features within one factor graph, with a separate readout operation in each channel summarizing a distinct aspect of the graph. For labeled data, we train the model using a supervised objective function while for unlabeled data, the model is trained with MI-based constraints for the input factorized graph and its corresponding features to ensure the disentanglement. Next we conduct MI maximization between the supervised and unsupervised models under each latent factor instead of in the whole feature space for disentangled consistency regularization. Compared with the existing works, this novel factor-wise MI estimation strategy can ensure the regularized factor is best pertinent to the aspect bridging supervised and unsupervised models for the current semi-supervised graph classification task. More importantly, we demonstrate that our framework can be formalized as a problem of maximizing the log-likelihood solved by Expectation Maximization (EM).

To summarize, we make the following contributions:
\begin{itemize}[leftmargin=*]
\item \textit{Conceptual:} We propose a novel disentangled representation learning framework for the semi-supervised graph classification task, which explicitly models the rationale factor that fits well between supervised and unsupervised learning models. And the proposed framework can be applied to other semi-supervised learning tasks.
% We propose to learn the disentangled representation for semi-supervised graph classification. To our best knowledge, this is the first attempt to explicitly model the rationale that fits well between supervised and unsupervised learning. And the proposed mechanism can be potentially generalized to other semi-supervised learning tasks.
\item \textit{Methodological:} We propose a graph disentangled encoder that produces factor-wise graph representations under decomposed factor graphs. Moreover, different MI-based constraints and consistency regularization are proposed to capture the characteristic differences and the connections between supervised and unsupervised learning models.
\item \textit{Experimental:} We perform extensive experiments on a range of public datasets to evaluate the performance of \method{}. Experimental results demonstrate the efficiency and outstanding interpretability of our proposed framework for semi-supervised graph classification tasks.
\end{itemize}

% The rest of our paper is organized as follows. We review the existing related works in Section~\ref{sec:related work}. In Section~\ref{sec:preliminary} and Section~\ref{sec:model}, we describe the prior knowledge and the details of our \method{}, respectively. Section~\ref{sec:experiment} offers extensive experimental results including quantitative comparisons,
% ablation studies, parameter sensitivity, visualization and case study. In the end, we give a conclusion in Section~\ref{sec:conclusion}.

% ch chFor l eannen eaxample, disentangle the information between the style and content of a given text.
% challenges since the complex formation of graphs. 
% 
\section{Related Work}
\label{sec:related work}

\subsection{Semi-supervised Graph Classification}
Semi-supervised learning has garnered increasing attention in recent years. It is associated with a paradigm that learns from both labeled and unlabeled data and encompasses two prominent techniques. The first line is the consistency regularization methods, which rely on the manifold or smoothness assumption and posit that minor perturbations of the data points should not affect the model's output.
% realistic perturbations of the data points should not change the output of the model. %~\cite{oliver2018realistic}
The most common structure is the Teacher-Student framework, which involves two models: the student and the teacher. This framework applies a consistency constraint between the predictions made by the student and the teacher models~\cite{sajjadi2016regularization, laine2017temporal, tarvainen2017mean}. 
% Among them, $\Pi$-Model~\cite{sajjadi2016regularization} generates teacher predictions based on perturbed models. Temporal Ensembling~\cite{laine2017temporal} further modifies $\Pi$-model by maintaining an Exponential Moving Average (EMA) of past epochs predictions, whereas Mean Teacher~\cite{tarvainen2017mean} utilizes the averaging of network parameters to generate a teacher model.
The second line is pseudo-labeling methods, which involve predicting the label distribution for unlabeled samples and selecting the most confident samples to provide additional guidance during the training process~\cite{sohn2020fixmatch, zhang2021flexmatch}. %~\cite{yang2022survey} subsequently

% For example, FixMatch~\cite{sohn2020fixmatch} applies weak and strong augmentations for the same input and uses the weakly-augmented version to generate hard pseudo labels for strongly-augmented version training. Based on curriculum learning, Flexmatch~\cite{zhang2021flexmatch} further improves FixMatch using class-specific flexible thresholds to select unlabeled data during the generation of pseudo labels. 
%ju2022ghnn, 
For graph-structured data, there are also attempts using consistency regularization~\cite{sun2020infograph, you2020graph, ju2022tgnn, yue2022label, ju2022ghnn}, pseudo-labeling~\cite{li2019semi, luo2022dualgraph, ju2022kgnn, luo2023towards} or other approaches~\cite{ng2018bayesian,ju2024focus} for semi-supervised learning. Typically, InfoGraph~\cite{sun2020infograph} learns supervised and unsupervised model respectively and estimates the MI between two models. GraphCL~\cite{you2020graph} and GLA~\cite{yue2022label} use contrastive learning to get graph representations, which are then used in the fine-tuning step for semi-supervised graph classification. DualGraph~\cite{luo2022dualgraph} and UGNN~\cite{luo2023towards} jointly learn the prediction and retrieval modules via posterior regularization during the pseudo-labeling process. GraphSpa~\cite{ju2024focus} employs an active learning approach to select informative graphs for semi-supervised model training. However, these approaches fail to disentangle the diverse underlying factors behind the graph data, making it challenging to identify suitable alignments between supervised and unsupervised models. 
% To tackle this limitation, \method{} employs to learn disentangled representation for semi-supervised graph classification.

\subsection{Disentangled Representation Learning}
The goal of disentangled representation learning is to acquire factorized representations capable of discerning and separating underlying latent factors within the observed data~\cite{bengio2013representation}. 
% By achieving such a disentangled representation, the model gains a deeper understanding of the intricate dependencies within the data, facilitating more insightful and meaningful analysis or manipulation of the underlying factors. 
Existing efforts in disentangled representation learning are primarily focused on computer vision~\cite{higgins2016beta, chen2016infogan}, %, yang2021causalvae
natural language processing~\cite{cheng2020improving, wang2022disencite} and recommendation~\cite{wang2022disenctr, qin2023disenpoi}. 

% For example, $\beta$-VAE~\cite{higgins2016beta} adds the hyper-parameter $\beta$ as the weight of the KL divergence for the Variational Autoencoders (VAEs) objective to balance the independence constraints
% and reconstruction accuracy. InfoGAN~\cite{chen2016infogan} decomposes the representation into noise and extra class code, and estimates the MI between the class code and corresponding data for controllable image generation. DisenCite~\cite{wang2022disencite} learns the semantics of different sections in the paper via disentangled representation for context-specific citation generation. MacridVAE~\cite{ma2019learning} models user behavior data to study hierarchical user intentions via macro and micro representation disentanglement. 
Recently, there has been a notable surge of interest in applying disentangled representation learning techniques to graph-structured data~\cite{ma2019disentangled, wang2020disenhan, yang2020factorizable, li2021disentangled}. DisenGCN~\cite{ma2019disentangled} learns disentangled node representations using a neighborhood routing mechanism, which partitions the node's neighborhood into several mutually exclusive parts. DisenHAN~\cite{wang2020disenhan} learns disentangled representation in Heterogeneous Information Network (HIN) by iteratively identifying the primary aspects of the relationships between node pairs and semantically propagating the corresponding information. FactorGCN~\cite{yang2020factorizable} takes the whole graph as input and produces block-wise
interpretable graph-level features for classification. DGCL~\cite{li2021disentangled} proposes a self-supervised graph disentangled representation framework via the contrastive learning method. UMGRL~\cite{mo2023disentangled} provides a self-supervised disentangled representation learning method for the multiplex graph to capture common and private graph information. Our work focuses on learning disentangled representation for the semi-supervised graph classification task. And our primary objective is to design a framework where a supervised model can learn from an unsupervised model through the latent semantic space that aligns well with the current graph classification task. 
\section{Problem Definition and Preliminaries}
\label{sec:preliminary}
% In this section, we first give the notation and formal definition of our problem, then we introduce some important concepts of mutual information and its estimation. 
\subsection{Problem Definition}
\textbf{Definition 1} (Graph). We define a graph as $G=(V, E, X)$, where $V$ and $E\in V\times V$ denote the node set and edge set of the graph, respectively. $X\in\mathbb{R}^{|V|\times d'}$ denote the node feature matrix, where each row $x_i\in\mathbb{R}^{d'}$ represents the initial feature of node $i$ and $d'$ is the node feature dimension. A graph is typically labeled if it contains a class label $y\in\{0,1\}^C$, where $C$ denotes the number of classes. Otherwise, the graph is considered unlabeled when the class label is unknown.
% A graph is unlabeled if its class label is unknown.

\textbf{Definition 2} (Semi-supervised Graph Classification). Given a set of graphs $\mathcal{G}=\{\mathcal{G}^L, \mathcal{G}^U\}$, where $\mathcal{G}^L=\{G_1,\dots, G_{|\mathcal{G}^L|}\}$ and $\mathcal{G}^U=\{G_{|\mathcal{G}^L|+1},\dots, G_{|\mathcal{G}^L|+|\mathcal{G}^U|}\}$ represent labeled and unlabeled graphs respectively. Let $\mathcal{Y}^L=\{y_1,\dots,y_{|\mathcal{G}^L|}\}$ represents the label corresponding to $\mathcal{G}^L$. Semi-supervised graph classification seeks to learn a prediction function that can assign class labels to the unlabeled graph data in $\mathcal{G}^U$ based on the class labels available in $\mathcal{G}^L$.
% Semi-supervised graph classification aims to learn a prediction function that determines the class label of the unlabeled graph data in $\mathcal{G}^U$ from the available class labels in $\mathcal{G}^L$.

\subsection{Preliminaries}
\textbf{Preliminary 1} (MI Estimation). MI is a key metric for assessing the dependence between two random variables, and it has been utilized across a diverse array of tasks~\cite{chen2016infogan, velickovic2019deep}. %hjelm2018learning,
However, estimation MI is intractable especially in high dimensional continuous settings. Since earlier non-parametric binning methods perform poor, a neural estimator MINE~\cite{belghazi2018mutual} formates MI between $x$ and $y$ as Kullback-Leibler (KL)-divergence between their joint distribution $p_{xy}$ and the product of their marginal distributions $p_x\otimes p_y$,
\begin{equation}
\label{eqt:1}
    I_{KL}(x,y) := D_{KL}(p_{xy}||p_x\otimes p_y),
\end{equation}
% of MI with dual representations 
and derive a lower bound of the KL-divergence for estimation. Following previous works~\cite{velickovic2019deep,sun2020infograph}, we can further rely on non-KL divergence (i.e., Jensen-Shannon divergences) as an alternative for the MI estimator.
\begin{equation}
\label{eqt:2}
\scalebox{0.98}{$
    I_{JS}(x,y) = \mathbb{E}_{p_{xy}}[-T(x,y)]-\mathbb{E}_{p_x\otimes p_y}[\log(1-T(x,y))],$}
\end{equation}
where $T(\cdot,\cdot)$ is a critic (or score) function approximated by a neural network. In this way, the loss can be seen as a standard binary cross-entropy (BCE) loss between positive samples from the joint distribution and negative samples from the product of the marginals.

% , where samples from the joint distribution are considered positive examples, and samples from the product of the marginals are considered negative examples.

% i.e., Donsker-Varadhan (DV)-representation:
% \begin{equation}
% \label{eqt:2}
% % \sup_{T:\Omega\leftarrow\mathbb{R}}
%     I_{DV}(x,y) = \mathbb{E}_{p_{xy}}(T(x,y))-\log(\mathbb{E}_{p_x\otimes p_y}\exp^{T(x,y)}),
% \end{equation}
% % on some compact domain $\Omega\leftarrow\mathbb{R}^d$ 
% where $T(\cdot,\cdot)$ is a critic (or score) function approximated by a neural network. 
% % to depict a differentiable and scalable lower bound. 

\textbf{Preliminary 2} (MI and Contrastive Learning). More recently, InfoNCE~\cite{oord2018representation} propose a Noise Contrastive Estimation (NCE~\cite{gutmann2010noise})-based lower bound for MI estimation, defined as:
\begin{equation}
\label{eqt:3}
% \sup_{T:\Omega\leftarrow\mathbb{R}}
    I_{NCE}(x,y) := \mathbb{E}\Bigl[\frac{1}{N}\sum_{i=1}^N\log\frac{\exp^{T(x_i, y_i)}}{\frac{1}{N}\sum_{j=1}^N \exp^{T(x_i, y_j)}}\Bigr],
\end{equation}
where $N$ samples $\{(x_i, y_i)\}_{i=1}^N$ drawn from $p_{xy}$ are used to compute the expectation. Contrastive learning encourages the model where representations between positive samples are pulled closer (e.g., graph representations corresponding to the same factor bridging supervised and unsupervised tasks) and representations among negative samples are pushed apart.

% , including self-supervised learning~\cite{hjelm2018learning,velickovic2019deep} and generative modeling~\cite{chen2016infogan}, etc
\section{The Proposed Model}
\label{sec:model}
% In this section, we first introduce the motivation and architecture of the proposed model DisenSemi. We then present the details of each component and the overall optimization for semi-supervised graph classification.
\subsection{Overview}
% The basic idea of semi-supervised learning is to 
For semi-supervised graph learning, the objective is to smooth the label information over graph data with regularization. %~\cite{yang2022survey}
And a straightforward way for the objective is to combine the purely supervised loss and the unsupervised objective function. Formally, previous semi-supervised learning frameworks try to minimize the following objective function:
\begin{equation}
\label{eqt:4}
   L_{total} = \sum_{i=1}^{|\mathcal{G}^L|}L_S(G_i, y_i)+\lambda\sum_{j=1}^{|\mathcal{G}|} L_U(G_j),
\end{equation}
where $L_S$ denotes the supervised loss, which quantifies the difference between the predictions of the supervised model and the actual labels of the graph data, $L_U$ denotes the unsupervised loss acting as a regularization term and $\lambda$ is the relative weight between two losses. 

\begin{figure}[t]
    \centering    
    \includegraphics[width=\linewidth]{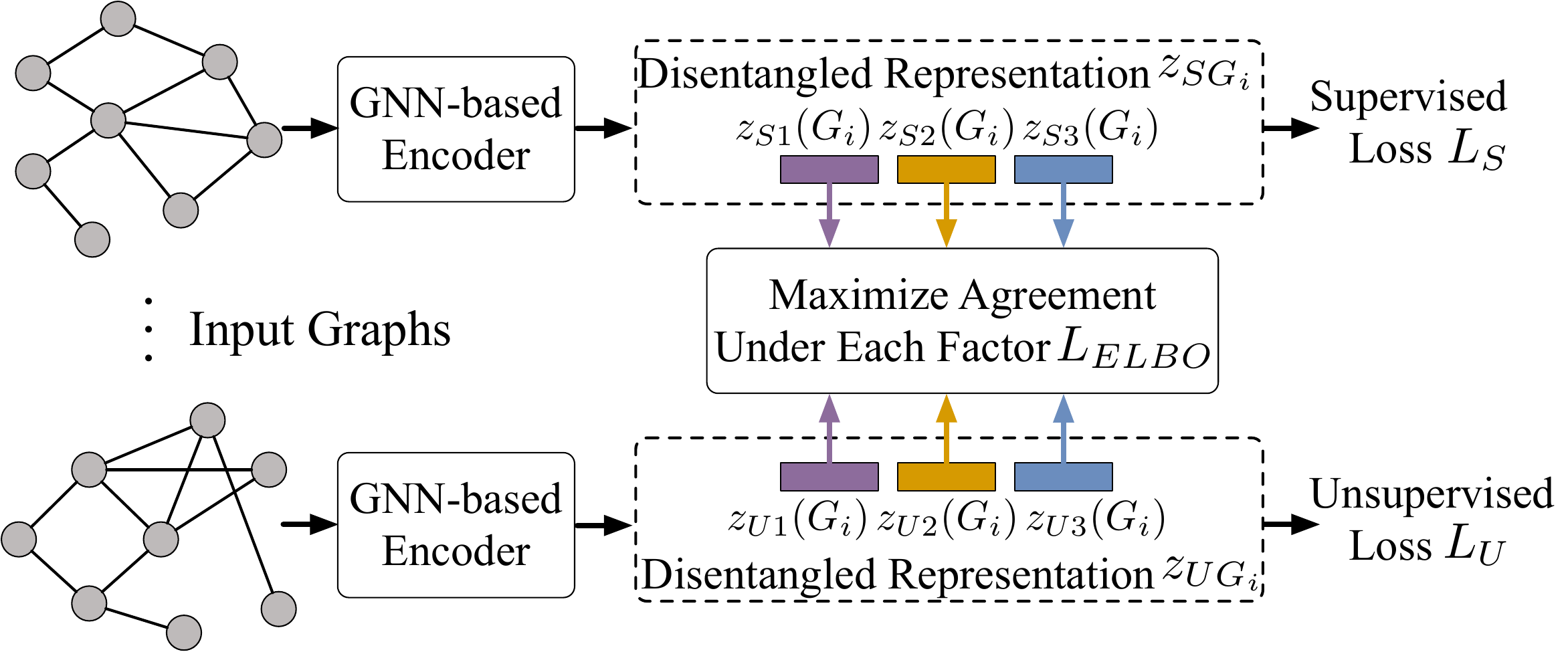}
    \caption{A schematic view of the \method{}, which consists of a supervised and an unsupervised model. We extract factor-wise graph representations from two models and maximize the agreement in a factor-wise manner.} 
    % }
    \label{fig:2}
\end{figure}

% However, we argue that supervised and unsupervised tasks have different optimization targets, which correspond to different semantic spaces of the graph data. Thus, we propose the \method{} for semi-supervised graph classification. The basic idea is to create a transfer framework that uses the knowledge from the well-aligned unsupervised model to enhance the supervised task.
However, we argue that supervised and unsupervised tasks have different optimization targets, which correspond to different semantic spaces of the graph data. Thus, we propose \method{} for semi-supervised graph classification. The basic idea is to explicitly infer the latent factors underlying a substantial amount of unlabeled graph data and transfer the well-aligned knowledge to enhance the supervised task in a factor-wise manner. As shown in Fig.~\ref{fig:2}, our framework is composed of a supervised model and an unsupervised model, where graph-level representations of both models are learned via the GNN-based encoder. Given the graph data, the GNN-based encoder first factorizes the input graph into several factor graphs, then gets graph representation by propagating and aggregating features on the corresponding factor graph. For the supervised model, we merge all the extracted factor graph representations to predict graph labels. For the unsupervised model, we introduce several MI-based constraints among factorized graphs and their corresponding extracted representations. Finally, a disentangled consistency regularization is conducted to explicitly identify the rationale between the two models and transfer the learned knowledge semantically via factor-wise MI maximization. The process is equivalent to a Variational EM algorithm maximizing the log-likelihood.

% we maximize the MI between factorized graph and its corresponding extracted representation in a self-supervised way, and minimize the MI between each representation to disentangle the graph representation effectively. 

% we deploy two GNN-based encoders for both supervised and self-supervised tasks, utilize the disentangled representations to explicitly identify the rationale between two tasks that best benefits the label prediction, and propose a factor-wise MI regularization to encourage positive knowledge transfer.

% The paper proposes the DisenSemi for semi-supervised graph classification, the basic idea is to design a transfer framework that leverages the knowledge self-supervised learned in unlabeled data to enhance the supervised task. However, previous methods usually either suffer from neglecting latent factor entanglement of graph data or mismatching semantics between supervised and self-supervised tasks. As a result, the transferred knowledge is biased and the method performance is not yet satisfactory.

% In order to explicitly enable transferring rationale knowledge that fits well between supervised and self-supervised model, DisenSemi learns the disentangled representation of graph data and conducts several factor-wise MI estimation strategies between two models. 

\subsection{GNN-based Encoder}
% GNNs have recently emerged as effective approaches for learning graph-structured data. Prevailing GNNs for learning graph representations are based on the neural message passing among nodes to update their features and utilize a readout function to get the graph-level representation. 
GNNs have recently become powerful tools for learning from graph-structured data. Current GNNs typically use neural message passing to update node features and apply a readout function to obtain graph-level representations. However, we argue that these methods are holistic and neglect the hidden factors stemming from different aspects. In fact, multiple relations between nodes are always mixed together and represented as one single edge, and these relations correspond to distinct discriminant aspects of the graph. Thus, given input graph $G_i$, we factorize it into $K$ factor graphs $\{G_{ik}\}_{k=1}^K$ to separately encode their features. Specifically, for each edge $e_m=\{x_u, x_v\}$ in edge set $E(G_i)=\{e_1,\dots,e_M\}$, where $M$ is the total edge of graph $G_i$, we project the node feature $x_v\in\mathbb{R}^{d'}$ with a transformation matrix $W\in\mathbb{R}^{d\times d'}$ and calculate the factor coefficients as:
\begin{equation}
\label{eqt:5}
    E_{uvk}(G_i)=\frac{1}{1+\exp^{-\Psi_k(Wx_u, Wx_v)}},
\end{equation}
where $\Psi_k$ is the function that computes the attention score for edge $e_m$ under factor 
$k$ with the features of nodes $u$ and $v$ as input, the function can be implemented via a multi-layer perceptron (MLP). Following that, the attention score is normalized to get $E_{uvk}(G_i)$, representing the coefficient of edge $e_m$ contained in the factor graph $G_{ik}$.

% takes the features of node $u$ and $v$ as input to compute the attention score of edge $e_m$ under factor $k$
We represent each factor graph by its own edge coefficient $E_{uvk}(G_i)$. By stacking $L$ message passing layers, the node embedding of each node $v$ in $G_{ik}$ is updated by recursively aggregating and combining its neighborhood features. Formally, the embedding of node $v$ at the $l$-th layer can be:
\begin{equation}
\label{eqt:6}
% \small
\begin{split}
    \!h_{N(v)}^{(l)}(G_i) &\!=\! \mathcal{A}_k^{(l)}\bigl(\{h_{uk}^{(l-1)}(G_i), \forall u\in N(v)\}, E_{uvk}(G_i)\bigr),\\
    \!h_{vk}^{(l)}(G_i) &\!=\! \mathcal{C}_k^{(l)}\bigl(h_{vk}^{(l-1)}(G_i), h_{N(v)}^{(l)}(G_i)\bigr), \\
    \!h_{vk}^{(0)}(G_i) &\!=\! W_kx_v,
\end{split}
\end{equation}
where $N(v)$ represents the neighborhood of node $v$, $\mathcal{A}_k^{(l)}$ represents the aggregation operation corresponding to the coefficient $E_{uvk}(G_i)$ as edge weight for factor graph $G_{ik}$, and $\mathcal{C}_k^{(l)}$ represents the combination operations at layer $l$. In this paper, we adopt GraphConv~\cite{morris2019weisfeiler} as our message passing layer due to its strong expressive power and initialize the node embedding of each factor graph via a factor-specific transformation matrix $W_K\in\mathbb{R}^{\frac{d}{K}\times d'}$.  

Finally, for each factor graph $G_{ik}$, the graph-level representations are derived by aggregating the embeddings of all nodes from the last $L$-th layer through a readout function. Formally,
\begin{equation}
\label{eqt:7}
   z_k(G_i)=\text{READOUT}(\{h_{vk}^{(L)}(G_i)\}_{v\in V(G_i)}),
\end{equation}
where $V(G_i)$ denotes the node set of $G_i$, $\text{READOUT}$ can be implemented as a simple permutation invariant function (i.e., mean function). By considering all the hidden factors of graph data, we can get a factor-wise graph representation, namely, $Z_{G_i}=[z_1(G_i),\dots,z_K(G_i)]$.
% $Z'=z_{G_1}||\dots||z_{G_k}, k\in[1,K]$, where $||$ is the concatenation operation.

% \subsection{Disentangled Semi-supervised Learning}
% For semi-supervised graph learning, the objective is to smooth the label information over graph data with regularization~\cite{}. And a straightforward way for the objective is to combine the purely supervised loss and the self-supervised objective function. Formally, previous semi-supervised learning framework try to minimize the following objective function:
% \begin{equation}
% \label{eqt:10}
%    L_{total} = \sum_{i=1}^{|\mathcal{G}^L|}L_S(G_i, y_i)+\lambda\sum_{j=1}^{|\mathcal{G}^L|+|\mathcal{G}^U|} L_U(G_j),
% \end{equation}
% where $L_S$ denotes the supervised loss that measures the discrepancy between the supervised model prediction and the real label of the graph data, $L_U$ denotes the self-supervised loss acting as a regularization term and $\lambda$ is the relative weight between two losses. 

% However, we argue that supervised task and self-supervised task have different optimization targets, which corresponding to different semantic space of the graph data. Thus, we deploy two GNN-based encoders for both supervised and self-supervised tasks, utilize the disentangled representations to explicitly identify the rationale between two tasks that best benefits the label prediction, and propose a factor-wise MI regularization to encourage positive knowledge transfer.

\subsection{Supervised Loss}
With the GNN-based encoder in the supervised task, a factor-wise representation $Z_{SG_i}=[z_{S1}(G_i),\dots, z_{SK}(G_i)]$ is extracted from labeled data, where different factor graphs will result in different features of the graph. Towards this end, we define $K$ learnable prototypes $\mathcal{C}=\{c_k\}_{k=1}^K$ to obtain the attention weight $\alpha_{G_{ik}}$ of factor $k$ for the input $G_i$:
\begin{equation}
\label{eqt:8}
   \alpha_k(G_i)=\text{Softmax}(\phi(z_{Sk}(G_i), c_k)),
\end{equation}
where $\phi$ denotes the similarity function (i.e., cosine similarity). After that, we weighted sum all latent factors to get the graph representation and fed it into a prediction model to get the output label. The process is shown in Fig.~\ref{fig:3}(b).
\begin{equation}
\label{eqt:9}
\begin{split}
   Z_{SG_i}' = \sum_{k=1}^K \alpha_k(G_i)\cdot z_{Sk}(G_i), \\
   \hat{y}_i = \text{Softmax}(\text{MLP}(Z_{SG_i}')),
\end{split}
\end{equation}
where we adopt a two-layer MLP to map the graph representation extracted from different factor graphs to label predictions. For the supervised loss, we adopt cross-entropy, defined as:
\begin{equation}
\label{eqt:10}
   L_S = -\sum_{c=1}^C y_{ic}\log\hat{y}_{ic},
\end{equation}
where $y_i\in\mathbb{R}^C$ is the label with the one-hot encoding for the graph $G_i$, $C$ corresponds to the total number of classes in the graph dataset. 
% Notice that we only calculate $L_S$ for labeled graphs.

\subsection{MI-based Representation Disentanglement}

\begin{figure*}[t]
    \centering   \includegraphics[width=0.97\textwidth]{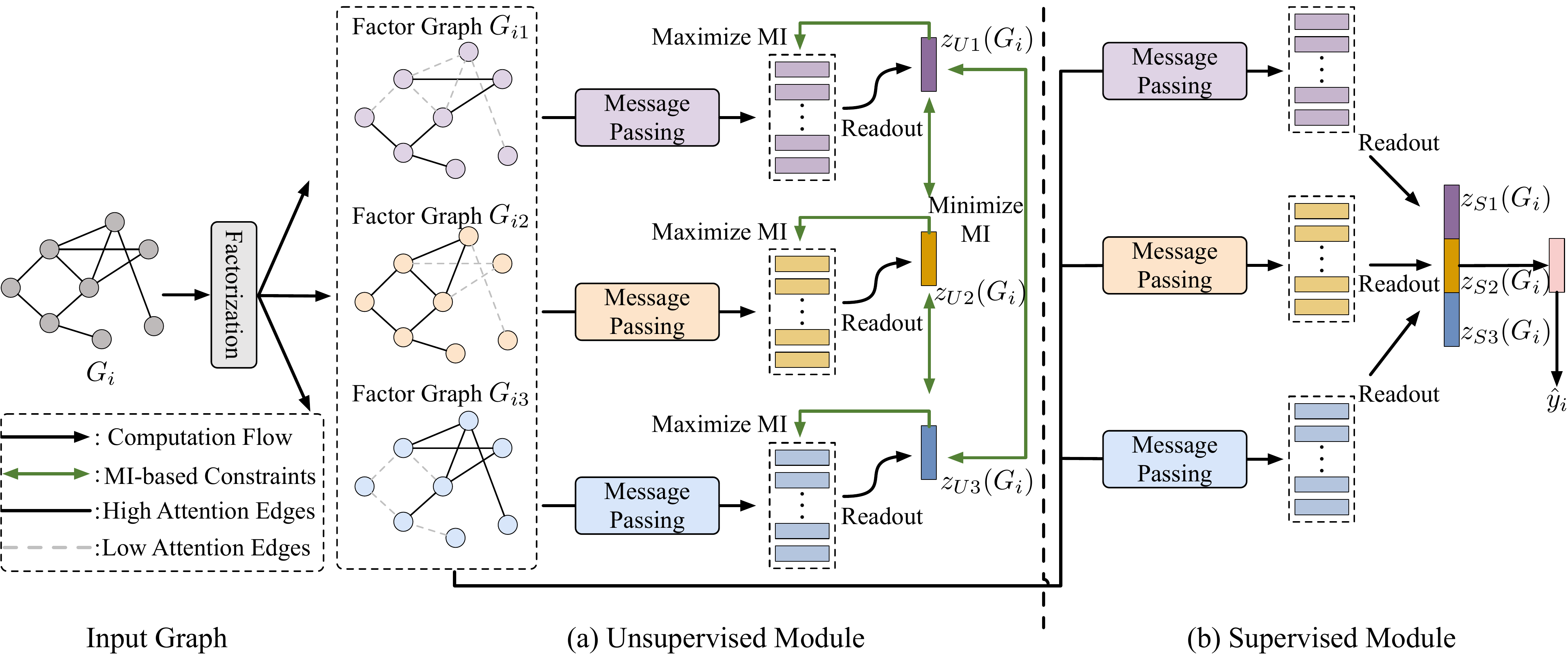}
    \caption{Illustration of the supervised and unsupervised training module. We assume that there are three aspects for the input graphs and factorize them into different factor graphs. For the unsupervised module, we maximize the intra-factor, and minimize the inter-factor MI to disentangle the graph representation effectively. Unlike MI-based constraints in the unsupervised module, we merge all extracted factor-wise graph representations to predict graph labels in the supervised module.}
% 
    
    % For unsupervised training module, we maximize the MI between factorized graph and its corresponding extracted representation (a.k.a., Intra-factor MI maximization), and minimize the MI between each representation (a.k.a., Inter-factor MI minimization) to disentangle the graph representation effectively.}
    \label{fig:3} 
\end{figure*}

For unsupervised learning, we can also get a factor-wise graph representation $Z_{UG_i}=[z_{U1}(G_i),\dots, z_{UK}(G_i)]$ via the GNN-based encoder. Since some of the decomposed factor graphs may exhibit similar structures without additional constraints, we propose several MI-based constraints to effectively disentangle the graph representation, as shown in Fig.~\ref{fig:3}(a).

\subsubsection{Intra-factor MI}
As each factor graph corresponds to an interpretable relation between nodes, we seek to obtain the discriminant representation that captures the global content of the graph. Inspired by the deep InfoMax~\cite{velickovic2019deep}, we maximize the MI between the decomposed factor graph and the corresponding graph-level representation. For factor $k$, we define the intra-factor MI on global-local pairs, maximizing the estimated MI over all the nodes in factor graph $G_{ik}$, which is shown as:
% Toward this end borrow the idea of deep InfoMax~\cite{velickovic2019deep} to
\begin{equation}
\label{eqt:11}
   L_{intra}^k = \frac{1}{|V|}\sum_{v\in V(G_i)} -I(z_{Uk}(G_i), h_{vk}^{(L)}(G_i)),
\end{equation}
where $I(z_{Uk}(G_i), h_{vk}^{(L)}(G_i))$ is the MI estimator, modeled by a discriminator $D_k$ and implemented through a neural network. We use the binary cross-entropy (BCE) loss between positive samples from the joint distribution and negative samples from the product of the marginals as the objective:
\begin{equation}
\label{eqt:12}
\scalebox{0.97}{$
\begin{aligned}
   \!-I(z_{Uk}(G_i), h_{vk}^{(L)}(G_i)) \!=\! \log \sigma(D_k(z_{Uk}(G_i), h_{vk}^{(L)}(G_i))) \\ + \mathbb{E}_{p\otimes\tilde{p}}\bigl[\log\bigl(1-\sigma(D_k(z_{Uk}(G_i), h_{vk}^{(L)}(\tilde{G_i})))\bigr)\bigr],
\end{aligned}$}
\end{equation}
where $h_{vk}^{(L)}(\tilde{G_i})$ is generated by the negative input graph $\tilde{G_i}$. We sample $\tilde{G_i}$ from a distribution $\tilde{p}=p$ to match the empirical distribution of the input space. In practice, we could sample negative samples by an explicit (stochastic) corruption function or directly sample negative graph instances in a batch.

% , which is identical to the empirical probability distribution of the input space.
\subsubsection{Inter-factor MI}
For graph representations $Z_{UG_i}$ extracted from different factor graphs, the MI between every two of them reaches its minimum value zero when $p(z_{Uk}(G_i), z_{Uk'}(G_i))=p(z_{Uk}(G_i))p(z_{Uk'}(G_i))$, which means that $z_{Uk}(G_i)$ and $z_{Uk'}(G_i)$ are independent to each other. Thus, MI minimization among them encourages the factor-wise graph representation to learn different aspects of information for the graph. Recently, several MI upper bounds~\cite{poole2019variational,cheng2020club} have been introduced for MI minimization. However, the estimation MI upper bound among these $K$ graph representations requires $K(K-1)$ times estimation, resulting in far more costs especially when $K$ is large. To alleviate this problem, as orthogonality is a specific instance of linear independence, we loosen the constraint of minimization MI to orthogonality and the method has also been demonstrated to be effective by many previous studies~\cite{liang2020attributed}:
% . The constraint has also been demonstrated to be effective by previous studies~\cite{wang2022disencite}
\begin{equation}
\label{eqt:13}
\begin{split}
   L_{inter} &= |{Z_{UG_i}'}^{\mathrm{T}}Z_{UG_i}'-I|, \\
   Z_{UG_i}' &= z_{U1}(G_i)||\dots||z_{UK}(G_i),     
\end{split}
\end{equation}
where $|\cdot|$ is the $L_1$ norm, $I$ is the identity matrix, $||$ is the concatenation operation.

\textit{Remark:} We have also tested other MI minimization methods, such as Contrastive Log-ratio Upper Bound (CLUB)~\cite{cheng2020club}, which estimates MI by the log-ratio difference between the conditional distribution of positive and negative pairs. And these methods do not significantly impact the performance. 

We maximize the MI between the factor graph and its corresponding graph representation to characterize each hidden factor while minimizing the MI among different factor graph representations to enforce representation disentanglement. Formally, unsupervised loss of DisenSemi can be:
% of the same factor
\begin{equation}
\label{eqt:14}
   L_U = \sum_{k=1}^K L_{intra}^k + L_{inter}.
\end{equation} 

\subsection{Disentangled Consistency Regularization}
For transferring the learned disentangled graph representations from the unsupervised encoder to the supervised encoder, we encourage representations extracted from two encoders with high MI when they correspond to the same factor of one graph data bridging supervised and unsupervised models. Specifically, we maximize NCE-based MI lower bound~\cite{oord2018representation}, which seeks to bring similar representations closer together while pushing dissimilar instances further apart. We treat each representation of a graph data instance as a distinct class and distinguish it from representations extracted from other graph data instances. Formally, given graph dataset $\mathcal{G}=\{G_i\}_{i=1}^{|\mathcal{G}^L|+|\mathcal{G}^U|}$, we assign each graph $G_i$ with a unique surrogate label $s_i=i$. Unlike the label $y_i$ contained in labeled graph, $s_i$ can be seen as the corresponding ID of the instance in the dataset $\mathcal{G}$. In this way, the MI can be defined as:
\begin{equation}
\label{eqt:15}
\begin{split}
   I_{NCE}(Z_{SG_i},Z_{UG_i}) &:= \log p(s_i|G_i)\\
    &= \log\mathbb{E}_{p(k|G_i)}[p(s_i|G_i,k)],
\end{split}
\end{equation}
where $p(s_i|G_i)$ corresponding to the identify discrimination over dataset, $p(k|G_i)$ denotes the probability distribution that $G_i$ is reflected via the $k$-th latent factor, namely, can be $\alpha_k(G_i)$. And $p(s_i|G_i,k)$ denotes to the discrimination subtask to identify the corresponding ID under the $k$-th latent factor, which can be defined as:
\begin{equation}
\label{eqt:16}
  p(s_i|G_i,k) = \frac{\exp\phi(z_{Sk}(G_i),z_{Uk}(G_i))}{\sum_{j=1}^{|\mathcal{G}|}\exp\phi(z_{Sk}(G_i),z_{Uk}(G_j))},
\end{equation}
where the subtask is over all graphs in the dataset. $\phi$ is the similarity function, adopting cosine similarity in our paper.

% . $p(k|G_i)$ is the probability distribution over $k$ latent factors, namely, can be $\alpha_{G_{ik}}$. Meanwhile, the identify discrimination subtask under $k$-th hidden factor
In practice, maximizing the log-likelihood across the entire graph dataset is difficult due to the latent factors. Therefore, we calculate the posterior distribution with Bayes' theorem, which can be defined as:
\begin{equation}
\label{eqt:17}
   p(k|G_i,s_i) = \frac{p(k|G_i)p(s_i|G_i,k)}{\sum_{k=1}^K p(k|G_i)p(s_i|G_i,k)}.
\end{equation}
Compared with prior distribution $p(k|G_i)$ that inferred only given the graph $G_i$, $p(k|G_i,s_i)$ is the probability that reflects how well the $k$-th latent factor aligns both supervised and unsupervised models when applied to the same graph data. However, computing the posterior probability is prohibitive due to the term $p(s_i|G_i,k)$ in Eq.~\ref{eqt:16}, which requires considering all instances in the dataset to compute the denominator. Therefore, we alternatively maximize the evidence lower bound (ELBO) of the log-likelihood, which is defined as:
\begin{equation}
\label{eqt:18}
\begin{split}
   \log p(s_i|G_i) \geq L_{ELBO} := \mathbb{E}_{q(k|G_i,s_i)}[\log p(s_i|G_i, k)] \\ - D_{KL}(q(k|G_i, s_i), p(k|G_i)),
\end{split}
\end{equation}
where $q(k|G_i,s_i)$ denotes a variational distribution that approximates the posterior probability $p(k|G_i,s_i)$. 
% And the derivation of the ELBO is provided in Appendix~A. 
Here, we formalize the variational distribution by:
\begin{equation}
\label{eqt:19}
   q(k|G_i,s_i) = \frac{p(k|G_i)\tilde{p}(s_i|G_i,k)}{\sum_{k=1}^K p(k|G_i)\tilde{p}(s_i|G_i,k)},
\end{equation}
where $\tilde{p}(s_i|G_i,k)$ can be defined with NT-Xent loss~\cite{chen2020simple} on a minibatch $\mathcal{B}\subseteq\mathcal{G}$ of graph data:
\begin{equation}
\label{eqt:20}
   \tilde{p}(s_i|G_i,k) = \frac{\exp\phi(z_{Sk}(G_i),z_{Uk}(G_i))}{\sum_{j\in\mathcal{B},j\neq i}^{|\mathcal{B}|}\exp\phi(z_{Sk}(G_i),z_{Uk}(G_j))}.
\end{equation}
Notice that the process is a variant of the Variational EM algorithm, where $q(k|G_i,s_i)$ is inferred during the E-step, and the ELBO is optimized during the M-step.

% \subsection{Evidence Lower Bound (ELBO)}
\subsection{Optimization and Complexity Analysis}

\subsubsection{Optimization}
We combine the supervised classification loss $L_S$, unsupervised loss $L_U$ as well as consistency regularization loss $L_{ELBO}$ together. And the overall objective for our semi-supervised graph classification is defined as:
\begin{equation}
\label{eqt:21}
   L_{total} = \sum_{i=1}^{|\mathcal{G}^L|}L_S + \lambda\sum_{j=1}^{|\mathcal{G}|} L_U -\gamma\sum_{j=1}^{|\mathcal{G}|} L_{ELBO},
\end{equation}
where $\lambda$ is the relative weight between losses, $\gamma$ is a tunable weight for consistency regularization loss. 

% . For each training iteration, we consider $B^L$ size of graphs from $\mathcal{G}^L$ to get $L_S$ and $B=B^L+B^U$ size of graphs to get $L_U$ and $L_{ELBO}$, while the negative sample is randomly chosen from the batch
% $B=B^L+B^U$ ($B^L$ from $\mathcal{G}^L$ and $B^U$ from $\mathcal{G}^U$)
\subsubsection{Complexity Analysis}
To ensure scalability with large-scale datasets, we leverage mini-batch with size $B=B^L+B^U$ to compute gradients. The computational consumption is mainly composed of three parts: (\romannumeral 1) the unsupervised module; (\romannumeral 2) the supervised module; (\romannumeral 3) the disentangled consistency regularization. Given the graph with an average number of nodes $|V|$ and edges $|E|$, the number of GNN layer and factor graphs is $L$ and $K$, and the representation dimension is $d$. For (\romannumeral 1) and (\romannumeral 2), the time complexity of the GNN-based encoder $O(|E|Ld)$. For (\romannumeral 1), the additional computational complexity of intra-factor and inter-factor MI constraints is $O(|V|d)$ and $O((K-1)d/2)$ respectively. For (\romannumeral 2), the additional computational complexity of the label prediction model is $O(d+(d/K)^2+Cd/K)$ with $C$ classes in the task. For (\romannumeral 3), we perform disentangled consistency regularization within minibatch, which takes $O(Bd)$ for each graph. To summarize, we have the overall complexity of \method{}, $O((2|E|L+|V|+(K+1)/2+d/K^2+C/K+B)d)$, which scales linearly w.r.t. the number of nodes and edges in each graph. Moreover, \method{} requires $K$ factor graphs, prototypes and coreresponding GNN-based encoder in both supervised and unsupervised module, the space complexity is $O((|V|+|E|)Ld+d+K|E|)$. 

\section{Experiment}
\label{sec:experiment}
We describe the experimental setup and present a series of extensive experiments conducted on ten real-world benchmark datasets to evaluate the performance of our proposed method. Our goal is to address the following research questions:

\begin{itemize}[leftmargin=*]
\item\textbf{RQ1:} How does our proposed method compare to other advanced models in semi-supervised graph classification?
\item\textbf{RQ2:} Are the key components of our method (GNN-based encoder, MI-based representation disentanglement, and consistency regularization) essential?
\item\textbf{RQ3:} How do different hyper-parameters affect the performance of our method?
\item\textbf{RQ4:} Can \method{} ensure the regularized factor pertinent to the aspect bridging supervised and unsupervised tasks semantically?
\end{itemize}

\subsection{Experimental Setups}
\subsubsection{Datasets and Baselines} We apply our model to ten public accessible graph classification benchmark datasets\footnote{https://ls11-www.cs.tu-dortmund.de/staff/morris/graphkerneldatasets;\\ https://ogb.stanford.edu/docs/graphprop/} following the previous works~\cite{sun2020infograph, yue2022label}, which includes four molecule datasets: MUTAG, PTC-MR, NCI1 and OGB-HIV, one bioinformatic dataset: PROTEINS and five social network datasets: IMDB-BINARY (IMDB-B), IMDB-MULTI (IMDB-M), REDDIT-BINARY (REDDIT-B), REDDIT-MULTI-5K (REDDIT-M5K) and COLLAB. Following recent works~\cite{sun2020infograph, you2020graph}, we adopt one-hot degree and centrality features instead when the datasets do not include node attributes. We compare our proposed \method{} with sixteen recent proposed baselines including seven traditional graph classification approaches: GK~\cite{shervashidze2009efficient}, SP~\cite{borgwardt2005shortest}, WL~\cite{shervashidze2011weisfeiler}, MLG~\cite{kondor2016multiscale}, DGK~\cite{yanardag2015deep}, Sub2Vec~\cite{adhikari2018sub2vec} and Graph2Vec~\cite{narayanan2017graph2vec} and nine GNN-based semi-supervised graph classification methods: MVGRL~\cite{hassani2020contrastive}, GraphCL~\cite{you2020graph}, JOAO~\cite{you2021graph}, DGCL~\cite{li2021disentangled}, RGCL~\cite{li2022let}, InfoGraph~\cite{sun2020infograph}, GLA~\cite{yue2022label}, TGNN~\cite{ju2022tgnn}, GraphSpa~\cite{ju2024focus}.
% the details are summarized in Appendix~C.
% The statistics of the datasets are summarized in Table~\ref{tab:1}.

\subsubsection{Evaluation Protocol} We assess the models using 10-fold cross-validation. Following the recent works~\cite{you2020graph, sun2020infograph, yue2022label}, we randomly shuffle and split each dataset into 10 parts. For each fold, one part is designated as the test set, another as the validation set, and the remaining parts are used for training. We then choose $30\%$ of the training graphs as labeled examples for each fold and apply semi-supervised learning. Results are reported as the average accuracy ($\%$) with standard deviation from ten repeated experiments.

\subsubsection{Implementation Details} 
% We implement our \method{} model in Pytorch. 
For all datasets, the embedding size of \method{} and other baselines (except GK, SP, WL, MLG) is fixed to 128. As the exact number of latent factors is unknown, we test different values for the number of factor graphs $K$ from the set $\{2^0,2^1,2^2,2^3,2^4\}$, corresponding embedding dimensions are $\{128,64,32,16,8\}$ per factor graph. We set three message passing layers for our method and the hyper-parameters $\lambda=0.001, \gamma=0.001$. 
% We optimize \method{} with Adam optimizer by setting the learning rate to 0.005 
The training is conducted for 200 epochs across all methods. For all the traditional graph classification approaches, we compute similarities between graphs and feed them into a downstream Support Vector Machines (SVM) classifier to evaluate the classification accuracy. For MVGRL, GraphCL, JOVO, DGCL and RGCL, we first pre-train the model with unlabeled graphs first and then fine-tune it with labeled graphs. For InfoGraph, GLA, TGNN, GraphSpa and our proposed \method{}, we integrate the pre-training and fine-tuning phases together. The source code of \method{} can be found at \url{https://github.com/jamesyifan/DisenSemi}.

\subsection{Performance Comparison (RQ1)}
\subsubsection{Overall Comparison}

\begin{table*}
% \begin{minipage}[t]{1\linewidth}
\setlength{\tabcolsep}{3.3pt}
\centering
\caption{Performance evaluation across ten benchmark datasets. We report the average prediction accuracy, along with the standard deviation from ten runs with varied random seeds (expressed in $\%$). Bolded results highlight the best performance.}
% \resizebox{0.8\textwidth}{!}{
\begin{tabular}{lcccccccccc}
% \begin{tabularx}{\textwidth}{lYYYYYYYYY}
\toprule
{ Methods} & {MUTAG} & {PTC-MR} & {NCI1} & {PROTEINS} & {IMDB-B} & {IMDB-M}  & {REDDIT-B}  & {REDDIT-M5K} & {COLLAB} & {OGB-HIV}\\
\midrule
GK & $82.9\pm2.1$ & $56.7\pm2.6$ & $64.7\pm0.2$ & $67.8\pm1.0$ & $57.9\pm0.7$ & $43.9\pm0.4$ & $68.4\pm0.5$ & $27.1\pm0.6$ & $62.3\pm0.6$ & $94.1\pm0.4$\\

SP & $84.5\pm2.5$ & $56.4\pm2.5$ & $69.8\pm0.3$ & $72.9\pm0.9$ & $51.9\pm0.6$ & $34.7\pm0.7$ & $76.0\pm1.1$ & $41.3\pm0.9$ & $58.6\pm0.3$ & $94.6\pm0.5$\\

WL & $86.8\pm2.1$ & $58.2\pm2.8$ & $76.2\pm0.3$ & $73.5\pm0.6$ & $69.2\pm1.0$ & $45.3\pm0.8$ & $67.7\pm1.1$ & $44.7\pm0.8$ & $74.3\pm0.2$ & $94.3\pm0.6$\\

MLG & $83.5\pm1.9$ & $58.8\pm2.2$ & $63.0\pm0.6$ & $72.3\pm1.1$ & $49.9\pm0.3$ & $38.8\pm0.6$ & $72.7\pm0.9$ & $35.3\pm0.8$ & $56.5\pm0.1$ & $94.8\pm0.5$\\

DGK & $82.9\pm2.6$ & $59.3\pm2.5$ & $78.8\pm0.6$ & $73.7\pm0.7$ & $69.9\pm0.7$ & $46.6\pm1.0$ & $78.7\pm0.6$ & $48.6\pm0.1$ & $77.4\pm0.2$ & $95.0\pm0.4$\\

Sub2Vec & $75.0\pm2.7$ & $57.6\pm2.2$ & $58.7\pm1.5$ & $63.8\pm1.2$ & $56.1\pm0.7$ & $35.0\pm1.1$ & $68.9\pm0.7$ & $34.0\pm0.6$ & $58.3\pm0.5$ & $94.4\pm0.6$\\

Graph2Vec & $74.9\pm1.4$ & $59.2\pm2.1$ & $70.6\pm0.3$ & $70.1\pm0.9$ & $63.7\pm1.1$ & $50.4\pm0.7$ & $76.6\pm1.0$ & - & - & -\\
\midrule

MVGRL & $88.8\pm0.7$ & $69.5\pm1.5$ & $72.6\pm0.5$ & $73.6\pm0.8$ & $74.5\pm0.7$ & $50.7\pm0.6$ & $90.0\pm0.2$ & $53.1\pm0.4$ & $77.8\pm0.2$ & $95.6\pm0.4$\\

GraphCL & $89.4\pm0.8$ & $70.0\pm1.6$ & $77.4\pm0.3$ & $76.1\pm0.6$ & $74.0\pm0.5$ & $50.3\pm0.5$ & $91.7\pm0.2$ & $57.2\pm0.6$ & $79.1\pm0.3$ & $95.8\pm0.5$\\

JOAO & $91.0\pm0.7$ & $70.6\pm1.3$ & $77.8\pm0.3$ & $76.7\pm0.8$ & $75.2\pm0.8$ & $50.8\pm0.5$ & $91.8\pm0.3$ & $\underline{57.4\pm0.2}$ & $79.6\pm0.1$ & $96.4\pm0.3$\\

DGCL & $90.5\pm0.9$ & $\underline{71.4\pm1.4}$ & $78.1\pm0.4$ & $76.8\pm0.7$ & $75.7\pm0.6$ & $\underline{51.7\pm0.7}$ & $91.5\pm0.5$ & $56.1\pm0.2$ & $\underline{80.3\pm0.4}$ & $96.4\pm0.2$\\

RGCL & $90.5\pm0.5$ & $71.2\pm1.2$ & $\underline{78.2\pm0.3}$ & $75.0\pm0.6$ & $\underline{75.9\pm0.6}$ & $50.1\pm0.4$ & $91.1\pm0.4$ & $56.4\pm0.3$ & $79.4\pm0.3$ & $96.3\pm0.4$\\

InfoGraph & $88.9\pm1.1$ & $70.6\pm1.4$ & $75.1\pm0.6$ & $74.9\pm0.8$ & $74.9\pm0.8$ & $51.5\pm0.6$ & $92.0\pm0.4$ & $56.9\pm0.5$ & $79.8\pm0.5$ & $95.6\pm0.3$\\

GLA & $\underline{91.6\pm1.0}$ & $70.6\pm1.7$ & $77.0\pm0.5$ & $\underline{77.1\pm0.6}$ & $75.8\pm0.6$ & $51.1\pm0.5$ & $\underline{92.2\pm0.3}$ & $56.7\pm0.2$ & $80.2\pm0.2$ & \underline{$96.8\pm0.3$}\\

TGNN & $91.3\pm0.6$ & $68.3\pm1.4$ & $75.2\pm0.7$ & $76.4\pm1.0$ & $75.1\pm0.7$ & $50.9\pm0.5$ & $91.5\pm0.5$ & $55.6\pm0.6$ & $77.6\pm0.4$ & $96.2\pm0.4$\\

GraphSpa & $91.5\pm0.9$ & $70.2\pm1.3$ & $77.2\pm0.5$ & $76.5\pm0.8$ & $75.4\pm0.6$ & $51.2\pm0.6$ & $91.9\pm0.6$ & $56.0\pm0.5$ & $78.9\pm0.3$ & $95.9\pm0.4$\\

\midrule 

\method{} & $\bf{92.6 \pm 0.6}$ & $\bf{72.4\pm1.1}$ & $\bf{78.9\pm0.4}$ & $\bf{78.4\pm0.5}$ & $\bf{76.7\pm0.7}$ & $\bf{52.5\pm0.5}$ & $\bf{93.2\pm0.3}$ & $\bf{57.6\pm0.2}$ & $\bf{81.5\pm0.3}$ &
$\bf{97.2\pm0.2}$\\ 
\bottomrule
% \end{tabularx}
\end{tabular}
% }
\label{tab:2}
% \end{minipage}
\end{table*}

Table~\ref{tab:2} provides the quantitative results of semi-supervised graph classification using a $30\%$ label ratio. From these results, we can draw the following conclusions.
\begin{itemize}[leftmargin=*]
\item Traditional graph classification approaches, which include graph kernel and unsupervised graph-level representation learning methods, perform worse than GNN-based semi-supervised graph classification methods. This indicates that hand-crafted features extracted by traditional methods suffer from poor generalization. Instead, GNN-based methods are more powerful to extract features from graph-structured data for the semi-supervised graph classification task.
\item Among GNN-based semi-supervised graph classification methods, DGCL and RGCL, which leverage disentangled representation for graph classification, always achieve better performance than other baselines on most datasets. It shows that explicitly considering the entanglement of factors for a graph helps to learn better representation for semi-supervised graph classification.
\item Our \method{}, outperforms all other models across the ten datasets, showcasing its superior effectiveness. The results indicate that the advancement beyond state-of-the-art methods stems not only from learning disentangled representation of the graph, but also from explicitly regularizing the rationale that fits well between supervised and unsupervised learning models for semi-supervised graph classification.
\end{itemize}

\subsubsection{Performance on Different Labeling Ratio}

\begin{figure*}
\centering
\begin{subfigure}{0.24\linewidth}
\includegraphics[width=\linewidth]{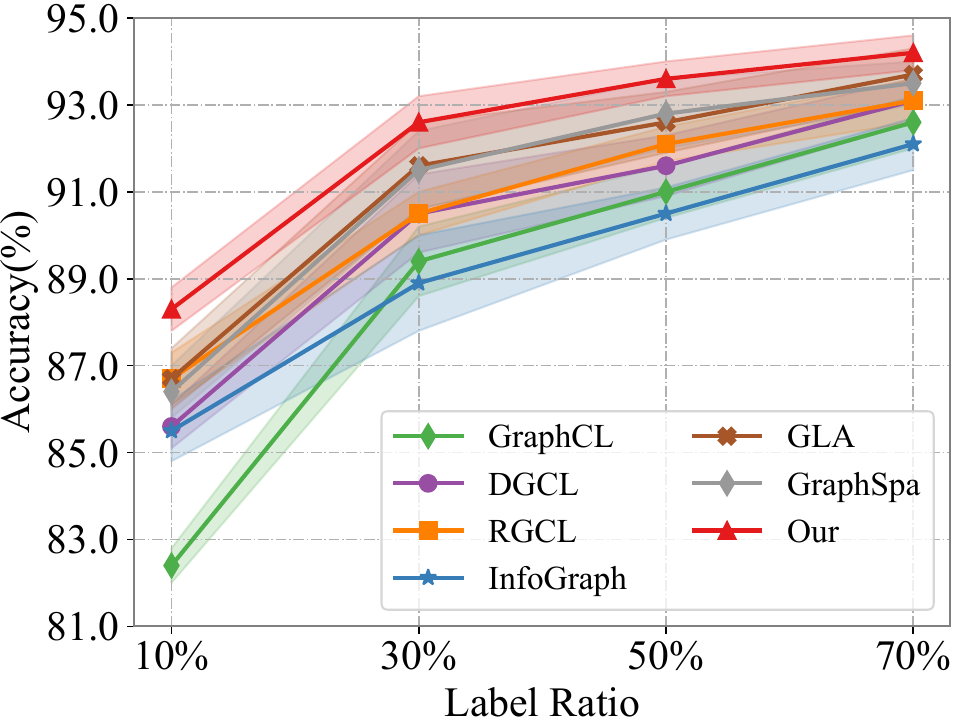}
\caption{MUTAG} \label{fig:4_a}
\end{subfigure}
\begin{subfigure}{0.24\linewidth}
\includegraphics[width=\linewidth]{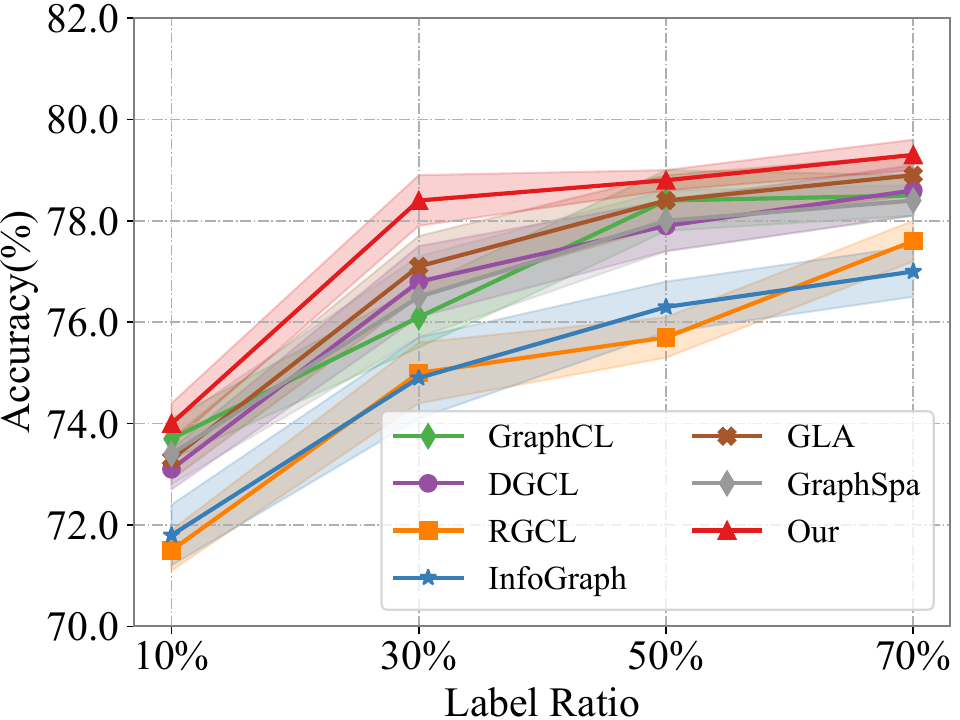}
\caption{PROTEINS} \label{fig:4_b}
\end{subfigure}
\begin{subfigure}{0.24\linewidth}
\includegraphics[width=\linewidth]{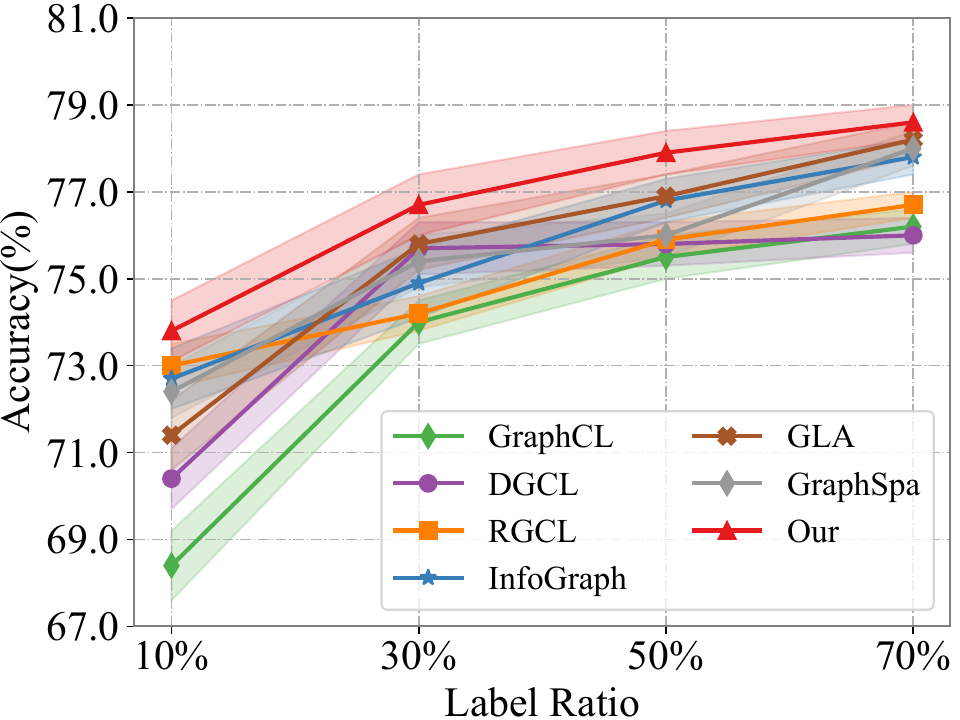}
\caption{IMDB-BINARY} \label{fig:4_c}
\end{subfigure}
\begin{subfigure}{0.24\linewidth}
\includegraphics[width=\linewidth]{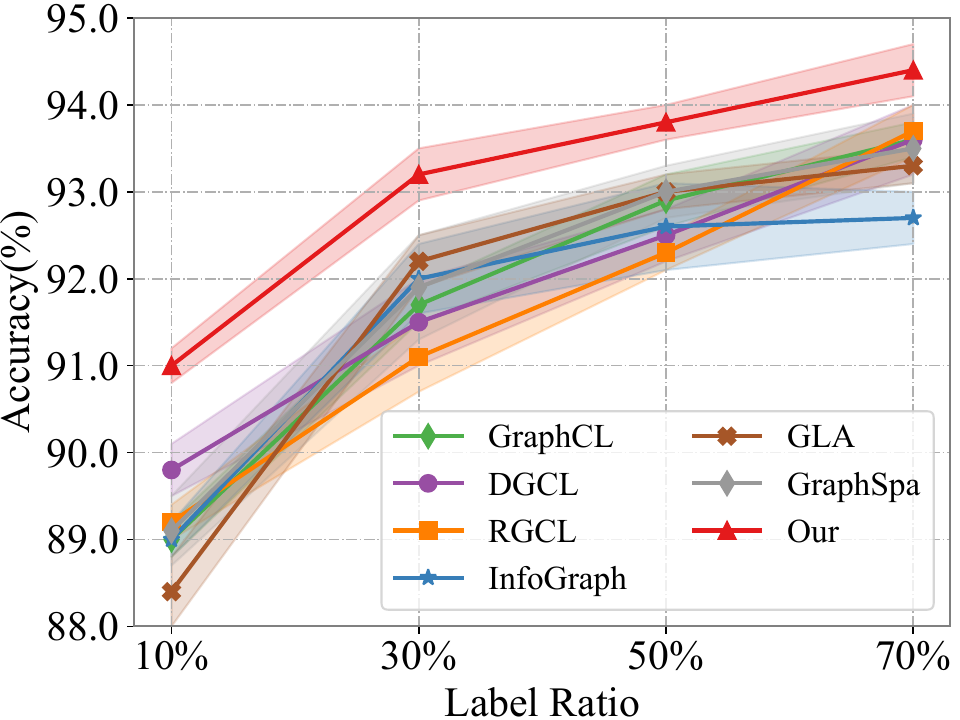}
\caption{REDDIT-BINARY} \label{fig:4_d}
\end{subfigure}
\caption{Performance comparison with different labeling ratios (i.e., $\%10$, $\%30$, $\%50$ and $\%70$) on four datasets (i.e., MUTAG, PROTEINS, IMDB-BINARY and REDDIT-BINARY).}
\label{fig:4}
\end{figure*}
% (i.e., MUTAG, PROTEINS, IMDB-BINARY and REDDIT-BINARY)
We vary the labeling ratio of training data to compare the performance of different models. Traditional methods are omitted due to their lack of competitive performance. Fig.~\ref{fig:4} illustrates the performance w.r.t. different labeling ratios of data on four datasets. We observe that:
\begin{itemize}[leftmargin=*]
\item As the quantity of labeled graph data increases, the performance of all models tends to improve. This demonstrates that the labeling ratio of training data is an important factor of the model. Meanwhile, the growth is not linear, which means unsupervised data also plays a key role to boost the model performance.
\item Disentangled graph representation model, i.e., DGCL and RGCL, though explicitly considering the entanglement of factors for graph, can not always guarantee the improvement over other baselines among different labeling ratios. This is attributed to the fact that supervised and unsupervised learning tasks for graph-structured data have different optimization targets. 
\item \method{} shows superior performance compared to other baselines for varying labeling ratios of data. Especially when the labeled data is scarce (e.g. less than $70\%$), the superiority of our model indicates that disentangled graph representation with consistency regularization can effectively improve the semi-supervised classification performance. 
\end{itemize}

\subsection{Ablation Studies (RQ2)}
To deeply understand our proposed \method{}, we conduct ablation studies over the key components of the model. 
% Particularly, we introduce a few model variants to study the effect of GNN-based encoder, MI-based representation disentanglement and consistency regularization. 

% Particularly, we investigate the effect of GNN-based encoder by exploring different encoder architecture, then we remove different parts of MI estimation to study the effect of MI-based representation disentanglement. At last, we introduce a few model variant to study the effect of consistency regularization
\subsubsection{Effect of GNN-based Encoder}
\begin{table*}
% \begin{minipage}[t]{1\linewidth}
\setlength{\tabcolsep}{3.15pt}
\centering
\caption{Ablation study on GNN-based encoder with different types of message passing layer and * means randomly removed $15\%$ edges from the original graph to obtain different factor graphs.}
% We report the mean together with the standard deviation of prediction accuracy over five runs (in $\%$) using different random seeds.
% \resizebox{0.8\textwidth}{!}{
\begin{tabular}{lcccccccccc}
% \begin{tabularx}{\textwidth}{lYYYYYYYY}
\toprule
{Methods} & {MUTAG} & {PTC-MR} & {NCI1} & {PROTEINS} & {IMDB-B} & {IMDB-M}  & {REDDIT-B}  & {REDDIT-M5K} & {COLLAB} & {OGB-HIV}\\
\midrule
SAGEConv* & $89.1\pm0.4$ & $68.4\pm1.2$ & $75.8\pm0.5$ & $74.7\pm0.5$ & $75.8\pm0.8$ & $51.9\pm0.5$ & $91.2\pm0.4$ & $56.5\pm0.3$ & $80.6\pm0.4$ & $95.6\pm0.5$\\

GATConv* & $89.6\pm0.3$ & $69.2\pm1.3$ & $75.2\pm0.6$ & $74.1\pm0.3$ & $75.6\pm0.8$ & $51.6\pm0.6$ & $92.1\pm0.4$ & $56.8\pm0.3$ & $80.4\pm0.3$ & $95.8\pm0.5$\\

GINConv* & $90.2\pm0.5$ & $68.9\pm1.1$ & $75.2\pm0.4$ & $74.7\pm0.4$ & $76.1\pm0.9$ & $52.2\pm0.5$ & $92.3\pm0.3$ & $56.9\pm0.2$ & $79.3\pm0.2$ & $96.0\pm0.3$\\

GraphConv* & $90.1\pm0.4$ & $69.1\pm1.0$ & $75.3\pm0.4$ & $74.6\pm0.5$ & $76.0\pm0.9$ & $51.4\pm0.6$ & $92.1\pm0.4$ & $56.7\pm0.3$ & $80.9\pm0.3$ & $95.8\pm0.4$\\ 

GCNConv & $90.6\pm0.5$ & $69.2\pm1.0$ & $75.5\pm0.6$ & $75.9\pm0.5$ & $76.0\pm0.8$ & $52.4\pm0.6$ & $92.4\pm0.2$ & $57.1\pm0.3$ & $81.2\pm0.4$ & $96.4\pm0.3$\\

GraphConv & $\bf{92.6\pm0.6}$ & $\bf{72.4\pm1.1}$ & $\bf{78.9\pm0.4}$ & $\bf{78.4\pm0.5}$ & $\bf{76.7\pm0.7}$ & $\bf{52.5\pm0.5}$ & $\bf{93.2\pm0.3}$ & $\bf{57.6\pm0.2}$ & $\bf{81.5\pm0.3}$ & $\bf{97.2\pm0.2}$\\ 
\bottomrule
% \end{tabularx}
\end{tabular}
% }
\label{tab:3}
% \end{minipage}
\end{table*}

In \method{}, we encode each factor graph via the proposed GNN with its own edge coefficient. To explore whether \method{} can derive benefits from the proposed GNN-based encoder, we compare five types of well-known GNN operators, namely SAGEConv~\cite{hamilton2017inductive}, GATConv~\cite{velivckovic2018graph}, GINConv~\cite{xu2018powerful}, GCNConv~\cite{kipf2017semi} and GraphConv~\cite{morris2019weisfeiler}. Notice that for SAGEConv*, GATConv* and GINConv*, we ignore the edge coefficient and randomly remove $15\%$ edges to obtain different factor graphs. For GraphConv*, we retain the edge coefficients and randomly remove edges separately. The performance of different GNN operators on all ten datasets is shown in Table~\ref{tab:3}. From the results, we can draw the following conclusions.
% \begin{itemize}[leftmargin=*]
% \item \textbf{SAGEConv}-It leverages GraphSAGE (with mean neighborhood aggregation)~\cite{hamilton2017inductive} as the encoder, which ignores its own edge coefficient and randomly removes $15\%$ edges to obtain different factor graphs.
% % treats each factor graph as having the same structure.
% \item \textbf{GATConv}-It encodes factor graphs with graph attention network (GAT)~\cite{velivckovic2018graph}, which ignores the edge coefficient and randomly removes edges to obtain factor graphs. 
% % treats each factor graph as having the same structure. 
% \item \textbf{GINConv}-It adopts graph isomorphism network (GIN)~\cite{hu2020strategies} as the encoder, which also ignores the edge coefficient and randomly removes edges to obtain factor graphs.
% % treats each graph as having the same structure. 
% \item \textbf{GCNConv}-It considers the edge coefficient of each factor graph and adopts graph convolutional network (GCN)~\cite{kipf2017semi} to encode each factor graph.
% \item \textbf{GraphConv}-It adopts a generalization GNN operator from previous work~\cite{morris2019weisfeiler} by considering the edge coefficient of each factor graph as well. 
% \end{itemize}
% The performance of different GNN operators on all nine datasets are shown in Table~\ref{tab:3}, we make the following observations from the results.
\begin{itemize}[leftmargin=*]
\item In most cases, the performance of GCNConv is better than other variants, which ignore the edge coefficient and randomly remove edges to obtain different factor graphs (i.e., SAGEConv*, GATConv*, GINConv*). This indicates that disentangling the graph with heterogeneous relations into different factor graphs can learn more effective and discriminative graph-level representation for the downstream classification task.
\item GraphConv* considers the high-order structure of the factor graph but still exhibits poor performance in cases where edges are randomly removed. Instead, GraphConv consistently outperforms other models when considering the edge coefficient. This outcome further validates the effectiveness of the learned edge coefficient within each factor graph for capturing higher-order graph structures.
% reason why we choose GraphConv as the backbone to encode each factor graph. 
% validates the effectiveness of our GNN-based encoder with strong representation ability. 
\end{itemize}
% especially when considering the edge coefficient instead of randomly removing edges. This also justifies the 

\subsubsection{Effect of MI-based Representation Disentanglement}
\begin{table*}
% \begin{minipage}[t]{1\linewidth}
\setlength{\tabcolsep}{3.1pt}
\centering
\caption{Ablation study on MI-based representation and disentanglement consistency regularization, w/o means remove the module from \method{} and Variant means different distribution of latent factors.}
% We report the mean together with the standard deviation of prediction accuracy over five runs (in $\%$) using different random seeds.
% \resizebox{0.8\textwidth}{!}{
\begin{tabular}{lcccccccccc}
% \begin{tabularx}{\textwidth}{lYYYYYYYY}
\toprule
{ Methods} & {MUTAG} & {PTC-MR} & {NCI1} & {PROTEINS} & {IMDB-B} & {IMDB-M}  & {REDDIT-B}  & {REDDIT-M5K} & {COLLAB} & {OGB-HIV}\\
\midrule
w/o Intra-MI & $90.8\pm0.5$ & $68.6\pm1.0$ & $74.9\pm0.8$ & $75.9\pm0.5$ & $75.8\pm0.8$ & $51.4\pm0.4$ & $92.6\pm0.2$ & $56.5\pm0.3$ & $80.8\pm0.4$ & ${95.9\pm0.4}$\\

w/o Inter-MI & $90.4\pm0.5$ & $69.2\pm1.2$ & $74.7\pm0.6$ & $76.2\pm0.4$ & $76.1\pm0.7$ & $52.3\pm0.6$ & $92.8\pm0.3$ & $56.2\pm0.4$ & $80.7\pm0.5$ & ${96.2\pm0.3}$\\

w/o MI & $89.3\pm0.4$ & $68.3\pm1.5$ & $74.1\pm0.7$ & $73.9\pm0.6$ & $75.3\pm0.8$ & $51.9\pm0.5$ & $91.6\pm0.3$ & $56.1\pm0.5$ & $80.1\pm0.4$ & ${95.6\pm0.5}$\\
\midrule
Variant 1 & $90.4\pm0.4$ & $69.8\pm1.0$ & $75.9\pm0.6$ & $74.1\pm0.5$ & $75.7\pm0.8$ & $51.8\pm0.5$ & $91.5\pm0.2$ & $56.3\pm0.5$ & $80.3\pm0.3$ & ${96.2\pm0.3}$\\

Variant 2 & $91.5\pm0.4$ & $68.6\pm0.9$ & $74.2\pm0.5$ & $74.5\pm0.4$ & $76.6\pm0.7$ & $52.3\pm0.7$ & $92.0\pm0.2$ & $56.4\pm0.4$ & $80.8\pm0.4$ & ${96.5\pm0.2}$\\
\midrule
\method{} & $\bf{92.6\pm0.6}$ & $\bf{72.4\pm1.1}$ & $\bf{78.9\pm0.4}$ & $\bf{78.4\pm0.5}$ & $\bf{76.7\pm0.7}$ & $\bf{52.5\pm0.5}$ & $\bf{93.2\pm0.3}$ & $\bf{57.6\pm0.2}$ & $\bf{81.5\pm0.3}$ & $\bf{97.2\pm0.2}$\\

\bottomrule
% \end{tabularx}
\end{tabular}
% }
\label{tab:4}
% \end{minipage}
\end{table*}

As disentangled graph representation is implemented via MI estimation in our \method{}, we investigate its impact on the performance with the following three variants: 
% (1) \textbf{w/o Intra-MI}-It removes intra-factor MI maximization on global-local pairs for unsupervised objective function (i.e., $L_U=L_{inter}$). (2) \textbf{w/o Inter-MI}-It removes inter-factor MI minimization between every two factor graph (i.e., $L_U = \sum_{k=1}^K L_{intra}^k$) for unsupervised objective function. (3) \textbf{w/o MI}-It removes the whole MI-based constraint for unsupervised objective function (i.e., w/o $L_U$). 
\begin{itemize}[leftmargin=*]
\item \textbf{w/o Intra-MI}-It removes the intra-factor MI maximization on global-local pairs for unsupervised objective function (i.e., $L_U=L_{inter}$). 
\item \textbf{w/o Inter-MI}-It removes the inter-factor MI minimization between every two factor graphs for unsupervised objective function (i.e., $L_U = \sum_{k=1}^K L_{intra}^k$).
\item \textbf{w/o MI}-It removes the whole MI-based constraint for unsupervised objective function (i.e., w/o $L_U$). 
\end{itemize}

Table~\ref{tab:4} presents the results of different model variants on all ten datasets and we can observe that the best results have been attained by considering both intra- and inter-MI estimation, which indicates that both intra-factor MI maximization on global-local pairs and inter-factor MI minimization between every two factor graphs play a key role to learn disentangled representation for semi-supervised graph classification. 
% \item Comparing with removing inter-factor MI minimization between every two factor graph (w/o Inter-MI), removing intra-factor MI maximization on global-local pairs (w/o Intra-MI) has more impact on the performance. This may be that 

\subsubsection{Effect of Consistency Regularization}
% \begin{table*}
% % \begin{minipage}[t]{1\linewidth}
% \setlength{\tabcolsep}{5pt}
% \centering
% \caption{Ablation studies on consistency regularization, Variant means different distribution of latent factors. We report the mean together with the standard deviation of prediction accuracy over five runs (in $\%$) using different random seeds.}
% % \resizebox{0.8\textwidth}{!}{
% \begin{tabular}{lccccccccc}
% % \begin{tabularx}{\textwidth}{lYYYYYYYY}
% \toprule
% { Methods} & {MUTAG} & {PTC-MR} & {NCI1} & {PROTEINS} & {IMDB-B} & {IMDB-M}  & {REDDIT-B}  & {REDDIT-M5K} & {COLLAB}\\
% \midrule
% Variant 1 & $90.4\pm0.4$ & $69.8\pm1.0$ & $75.9\pm0.6$ & $74.1\pm0.5$ & $75.7\pm0.8$ & $51.8\pm0.5$ & $91.5\pm0.2$ & $56.3\pm0.5$ & $80.3\pm0.3$\\

% Variant 2 & $91.5\pm0.4$ & $68.6\pm0.9$ & $74.2\pm0.5$ & $74.5\pm0.4$ & $76.6\pm0.7$ & $52.3\pm0.7$ & $92.0\pm0.2$ & $56.4\pm0.4$ & $80.8\pm0.4$\\

% \method{} & $\bf{92.6\pm0.6}$ & $\bf{72.4\pm1.1}$ & $\bf{78.9\pm0.4}$ & $\bf{78.4\pm0.5}$ & $\bf{76.7\pm0.7}$ & $\bf{52.5\pm0.5}$ & $\bf{93.2\pm0.3}$ & $\bf{57.6\pm0.2}$ & $\bf{81.5\pm0.3}$\\ 
% \bottomrule
% % \end{tabularx}
% \end{tabular}
% % }
% \label{tab:5}
% % \end{minipage}
% \end{table*}

Disentangled consistency regularization helps to transfer the learned disentangled graph representations from the unsupervised encoder to the supervised encoder. 
To further investigate how the consistency regularization facilitates the model performance, we evaluate \method{} against the following two variants:
\begin{itemize}[leftmargin=*]
\item \textbf{Variant 1}-It set $p(k|G_i)=1/K$, a uniform distribution over $K$ latent factors, which means each $k$-th latent factors reflected in $G_i$ is same.
\item \textbf{Variant 2}-It set $p(k|G_i)$ with a random distribution over $K$ latent factors. We implement it by randomly choosing a latent factor reflected in $G_i$.
\end{itemize}

Table~\ref{tab:4} displays the outcomes for \method{} and its variants. The observations are as follows:
\begin{itemize}[leftmargin=*]
\item The performance decreases when $p(k|G_i)$ of the model are set with uniform and random distribution in Variant 1 and Variant 2 respectively, which demonstrates that inferring the latent factor of a graph to explicitly identify and regularize the rationale between supervised and unsupervised learning model is important.
\item Variant 1 performs worse than variant 2 in most datasets. This may be that setting $p(k|G_i)$ with a random distribution can also specify a rationale between supervised and unsupervised learning models. Instead, inferring the latent factor of a graph uniformly might introduce rationale’s complement to the representation learning.
\end{itemize}

\subsection{Parameter Sensitivity (RQ3)}
We also analyze how the proposed \method{} responds to different hyper-parameter settings.
% We also examine the sensitivity of the proposed \method{} to various hyper-parameters.
% Specifically, we investigate the effect of varying numbers of factor graphs and message passing layers in our framework.

\subsubsection{Effect of the Number of Factor Graphs}

\begin{figure*}
\centering
\begin{subfigure}{0.24\linewidth}
\includegraphics[width=\linewidth]{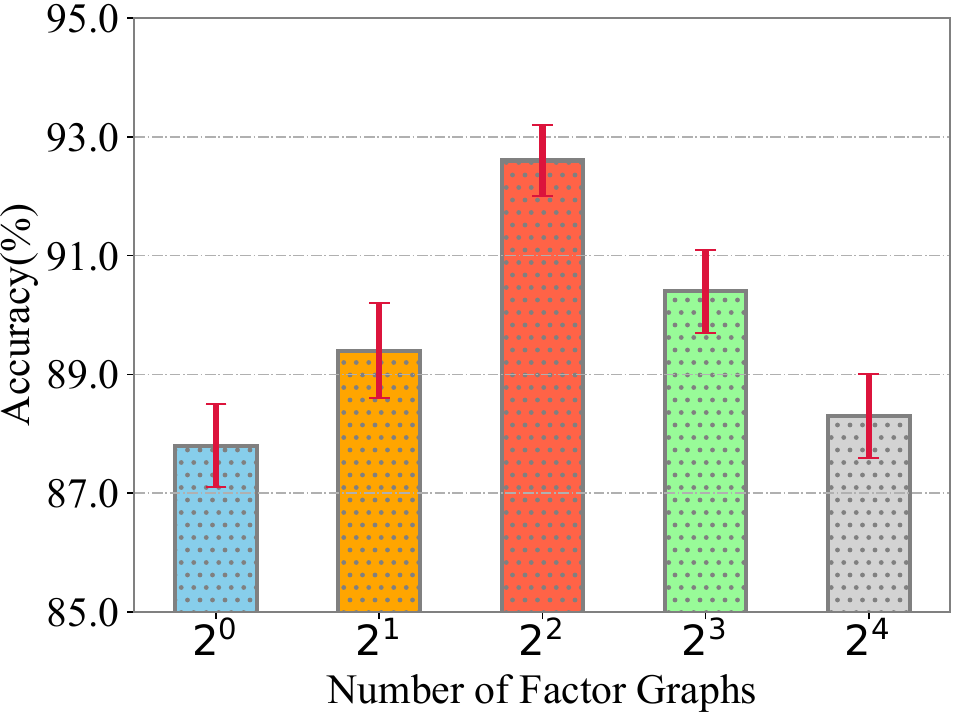}
\caption{MUTAG} \label{fig:5_a}
\end{subfigure}
\begin{subfigure}{0.24\linewidth}
\includegraphics[width=\linewidth]{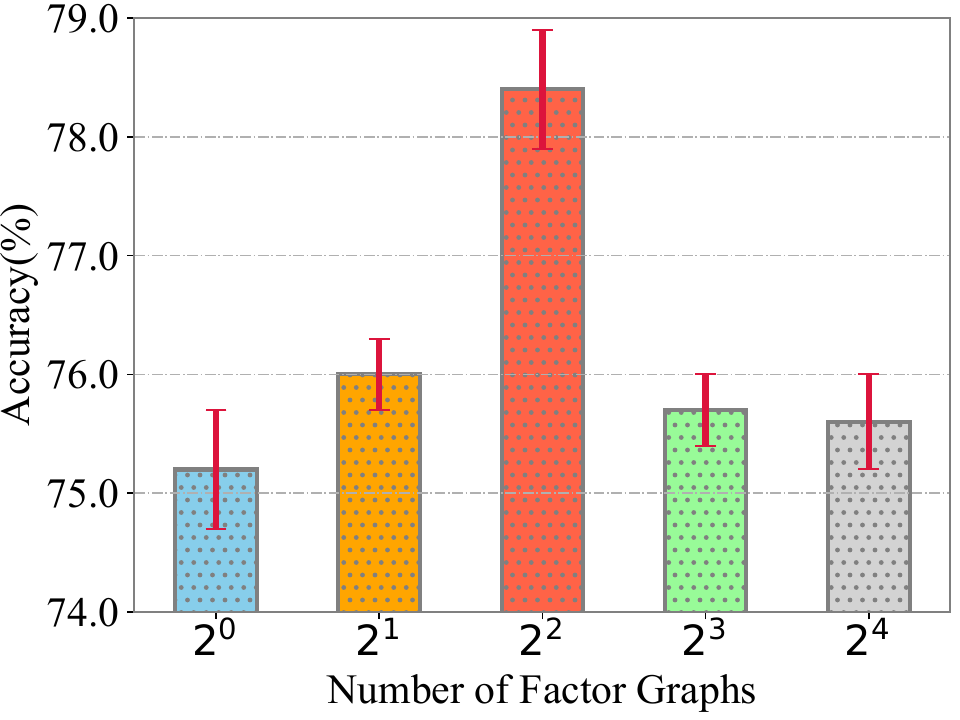}
\caption{PROTEINS} \label{fig:5_b}
\end{subfigure}
\begin{subfigure}{0.24\linewidth}
\includegraphics[width=\linewidth]{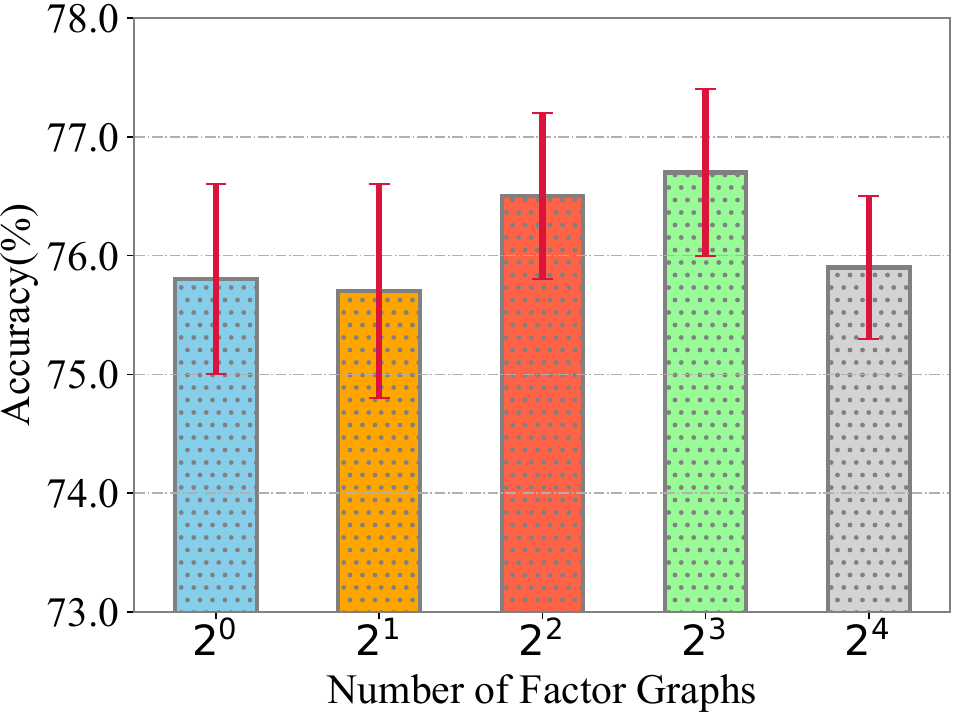}
\caption{IMDB-BINARY} \label{fig:5_c}
\end{subfigure}
\begin{subfigure}{0.24\linewidth}
\includegraphics[width=\linewidth]{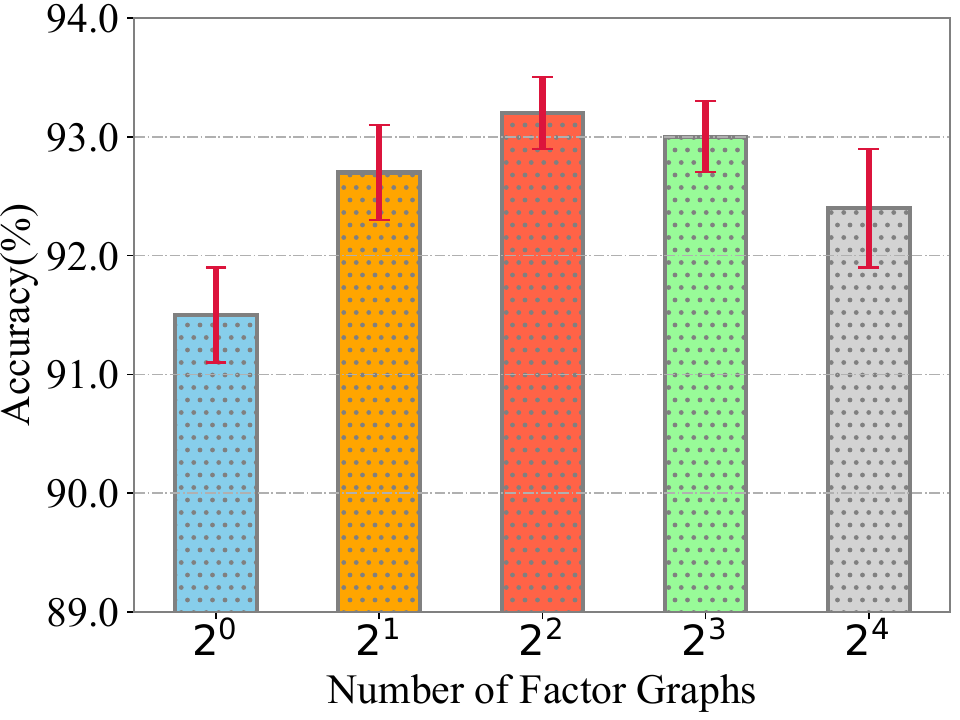}
\caption{REDDIT-BINARY} \label{fig:5_d}
\end{subfigure}
\caption{Performance w.r.t. different numbers of factor graphs in four datasets (i.e., MUTAG, PROTEINS, IMDB-BINARY and REDDIT-BINARY).}
\label{fig:5}
\end{figure*}

To assess the advantages of disentangled representation for \method{}, we evaluate the model's performance with different quantities of factor graphs. Specifically, we explore the number of factor graphs in the set $\{2^0,2^1,\dots,2^4\}$. Figure~\ref{fig:5} presents a summary of the experimental outcomes across four datasets, revealing the following insights:
\begin{itemize}[leftmargin=*]
\item When the number of factor graphs $K=1$, the model can be degraded into an entangled representation-based semi-supervised graph classification model with poor performance. This indicates that explicitly modeling multiple aspects of discriminant features for the graph can greatly facilitate the model performance.
\item Increasing the number of factor graphs can substantially enhance the model performance. \method{} achieves the best performance at $K=4$ in MUTAG, PROTEINS and REDDIT-B and $K=8$ in IMDB-B, which represents the optimal aspects of discriminant features for the graph.
\item Nonetheless, if the number of factor graphs is excessively large (e.g., $K\geq16$), the model’s performance tends to decline gradually. This degradation may result from employing an overly complex semantic structure for the graph.
\end{itemize}

\subsubsection{Effect of the Number of Message Passing Layers}

\begin{figure*}
\centering
\begin{subfigure}{0.24\linewidth}
\includegraphics[width=\linewidth]{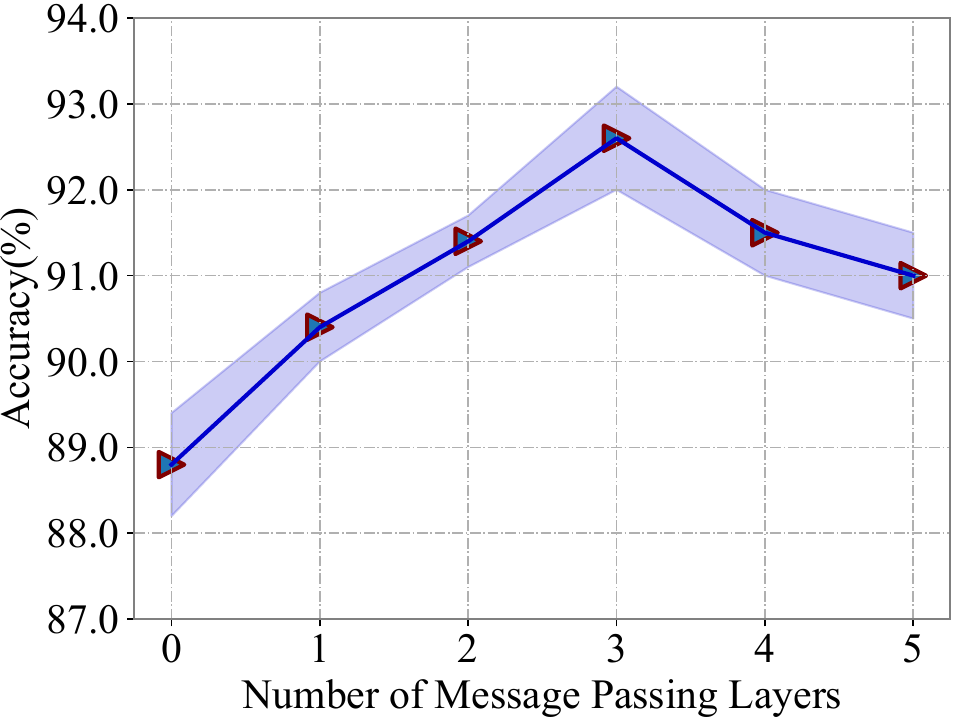}
\caption{MUTAG} \label{fig:6_a}
\end{subfigure}
\begin{subfigure}{0.24\linewidth}
\includegraphics[width=\linewidth]{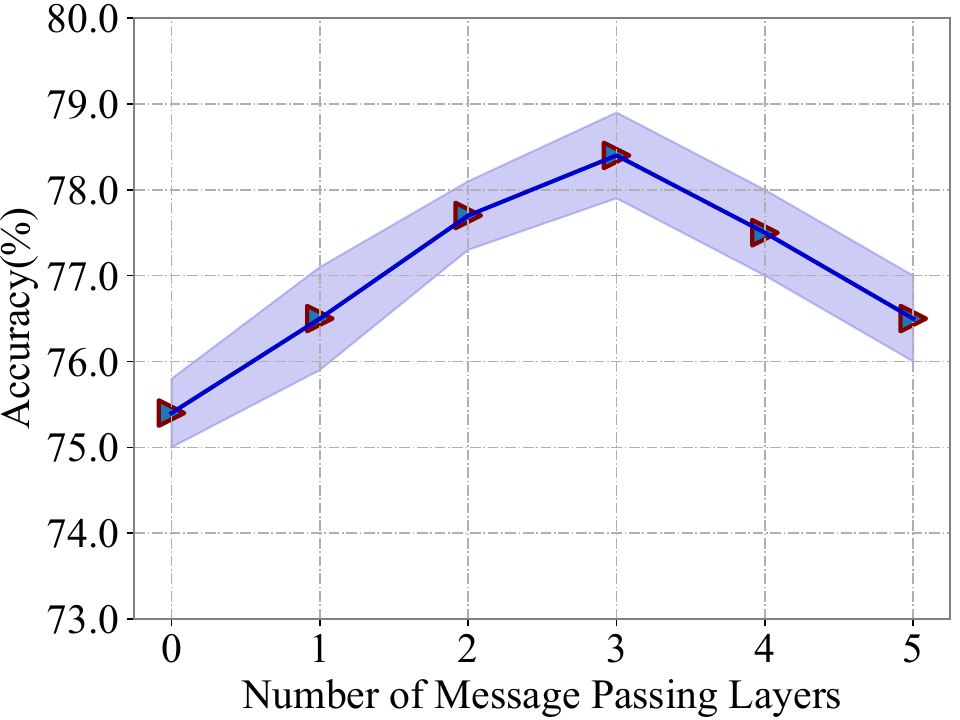}
\caption{PROTEINS} \label{fig:6_b}
\end{subfigure}
\begin{subfigure}{0.24\linewidth}
\includegraphics[width=\linewidth]{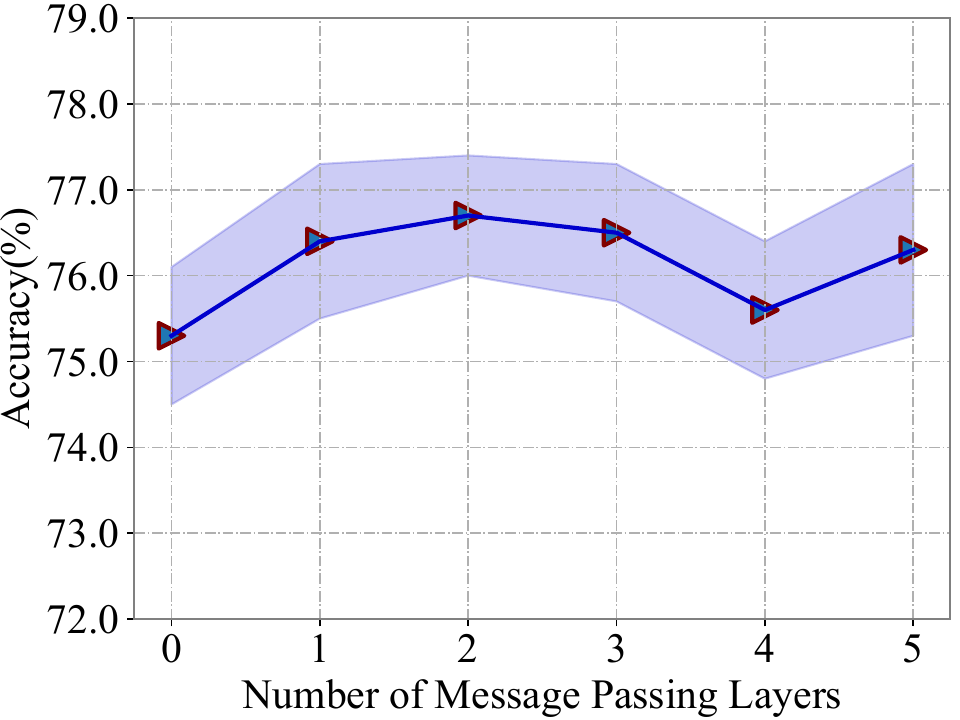}
\caption{IMDB-BINARY} \label{fig:6_c}
\end{subfigure}
\begin{subfigure}{0.24\linewidth}
\includegraphics[width=\linewidth]{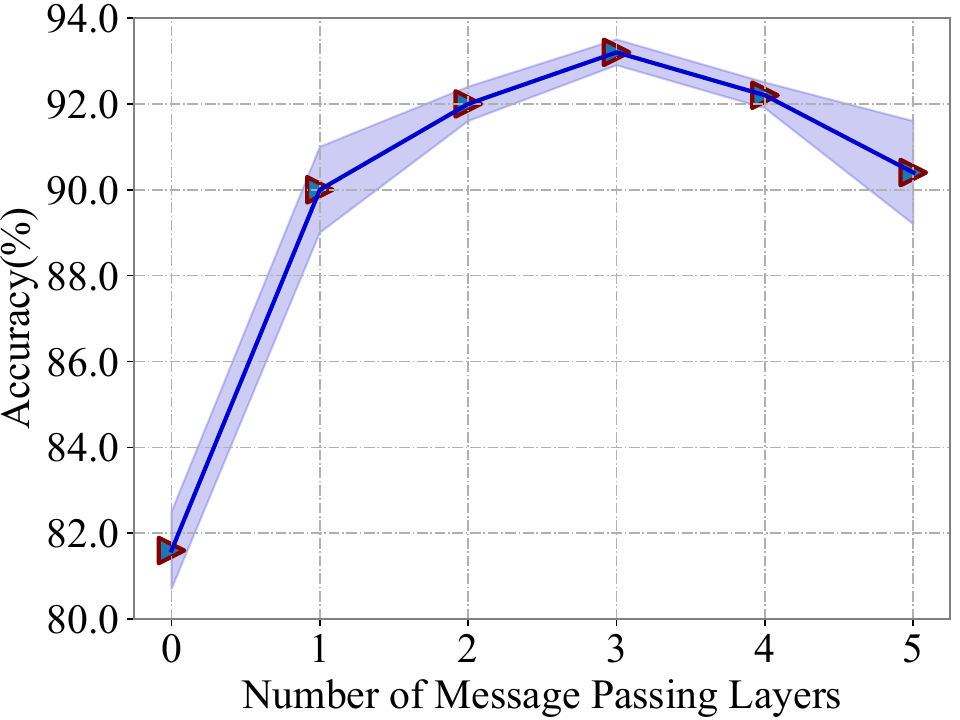}
\caption{REDDIT-BINARY} \label{fig:6_d}
\end{subfigure}
\caption{Performance w.r.t. different numbers of message passing layers in four datasets (i.e., MUTAG, PROTEINS, IMDB-BINARY and REDDIT-BINARY).}
\label{fig:6}
\end{figure*}

\begin{figure}
\centering
\begin{subfigure}{0.49\linewidth}
\includegraphics[width=\linewidth]{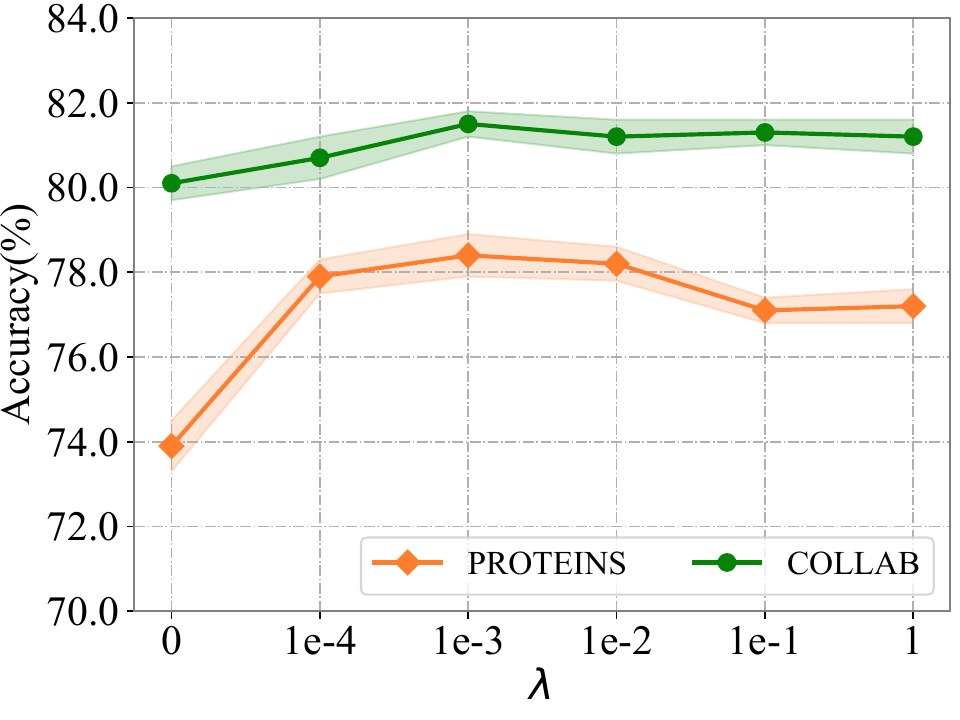}
\caption{Hyper-parameter $\lambda$}\label{fig:10_a}
\end{subfigure}
\begin{subfigure}{0.49\linewidth}
\includegraphics[width=\linewidth]{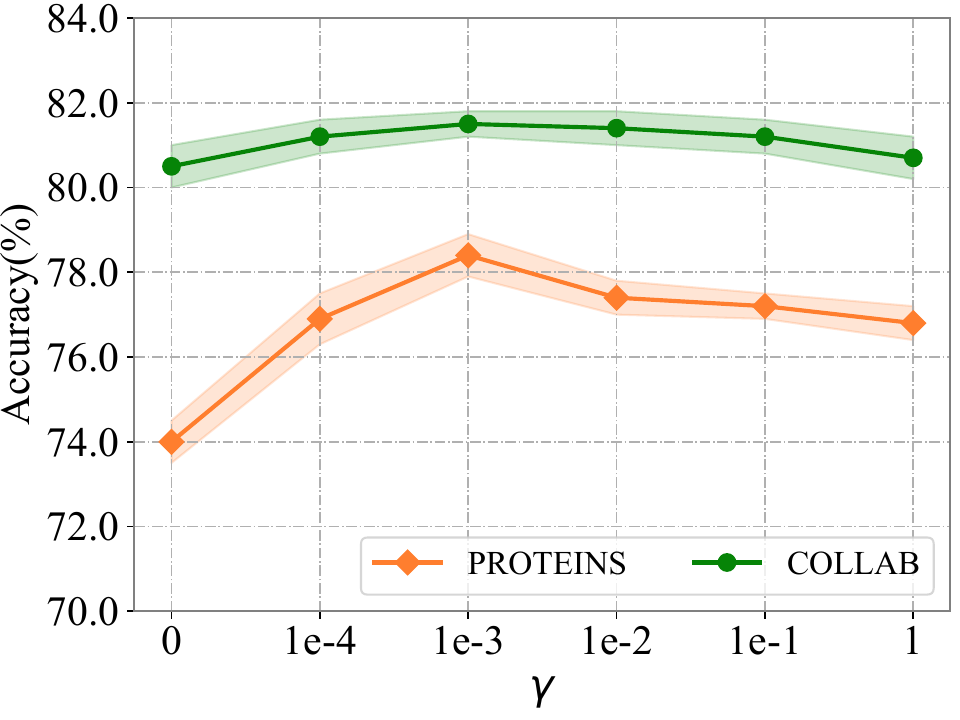}
\caption{Hyper-parameter $\gamma$}\label{fig:10_b}
\end{subfigure}
\caption{Performance w.r.t. different relative weight between losses in PROTEINS and COLLAB.}
\label{fig:10}
\end{figure}

To examine the impact of depth on performance for each disentangled factor graph, we test the \method{} model with a range of message passing layers from $[0,5]$. Figure~\ref{fig:6} illustrates the results of these experiments across four datasets—MUTAG, PROTEINS, IMDB-BINARY, and REDDIT-BINARY. The findings are as follows:
\begin{itemize}[leftmargin=*]
\item When the number of message passing layers $L=0$, the model only takes a linear
transformation into consideration and suffers from the degenerating issue. Hence, the result verifies the rationality and effectiveness of GNNs for learning graph-level representation.
\item Increasing the number of message passing layers incorporates information from higher-order neighbors and deepens the model. Evidently, a model with two layers performs better than one with only one layer. It illustrates the importance of messages passing between nodes to capture the structure of each factor graph. 
\item Too many message passing layers may hurt the model performance. When stacking more than three message passing layers, the model may introduce noise and suffer from the over-smoothing issue.
\end{itemize}

\subsubsection{Effect of Relative Weight between Losses}
We investigate the sensitivity of relative weight between losses, namely $\lambda$ and $\gamma$. In particular, we adjust the value of $\lambda$ and $\gamma$ from $0$ to $1$ on PROTEINS and COLLAB. Figure~\ref{fig:10} demonstrates that both the unsupervised module and disentangled consistency regularization play an important role in the objective and there is an evident performance drop when $\lambda,\gamma=0$. As $\lambda$ and $\gamma$ get larger, the performance is relatively stable, and we can get the best results when $\lambda,\gamma=1e-3$.

\subsection{Visualization and Case Study (RQ4)}
To further investigate how the disentangled representation facilitates the semi-supervised graph classification task, we conduct two qualitative assessments (visualization and case study) of the proposed \method{} and baselines.
% , including the visualization of the correlations between the elements of the learned representations and the case study of the learned factor graph. 
% \begin{figure*}
% \centering
% \begin{subfigure}{0.32\linewidth}
% \includegraphics[width=\linewidth]{figure/fig7_a.pdf}
% \caption{GraphCL} \label{fig:7_a}
% \end{subfigure}
% \begin{subfigure}{0.32\linewidth}
% \includegraphics[width=\linewidth]{figure/fig7_b.pdf}
% \caption{DGCL} \label{fig:7_b}
% \end{subfigure}
% \begin{subfigure}{0.32\linewidth}
% \includegraphics[width=\linewidth]{figure/fig7_c.pdf}
% \caption{\method{}} \label{fig:7_c}
% \end{subfigure}
% % \begin{subfigure}{0.32\linewidth}
% % \includegraphics[width=\linewidth]{Figure/fig8_c.pdf}
% % \caption{IMDB-BINARY} \label{fig:8_c}
% % \end{subfigure}
% \caption{Representation correlation analysis, where representations are obtained from the test split in MUTAG dataset. We use the fine-tuned model of GraphCL and DGCL, and supervised model of our proposed \method{} to get the representation. }
% \label{fig:7}
% \end{figure*}

\begin{figure*}
\centering
\begin{subfigure}{0.245\linewidth}
\includegraphics[width=\linewidth]{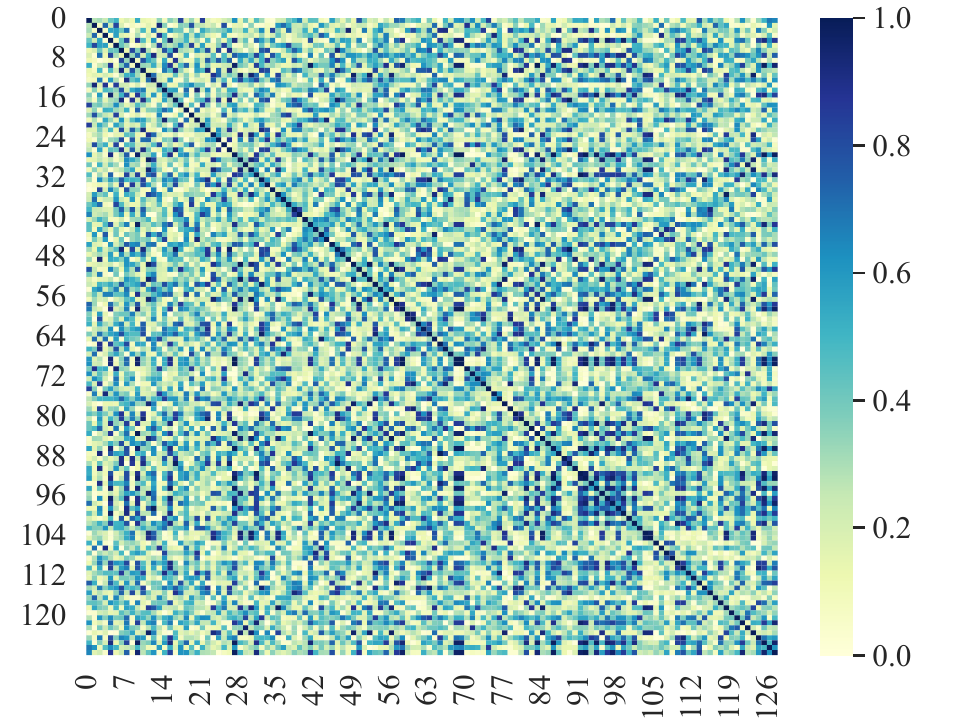}
\caption{GraphCL (MUTAG)} \label{fig:7_a}
\end{subfigure}
\begin{subfigure}{0.245\linewidth}
\includegraphics[width=\linewidth]{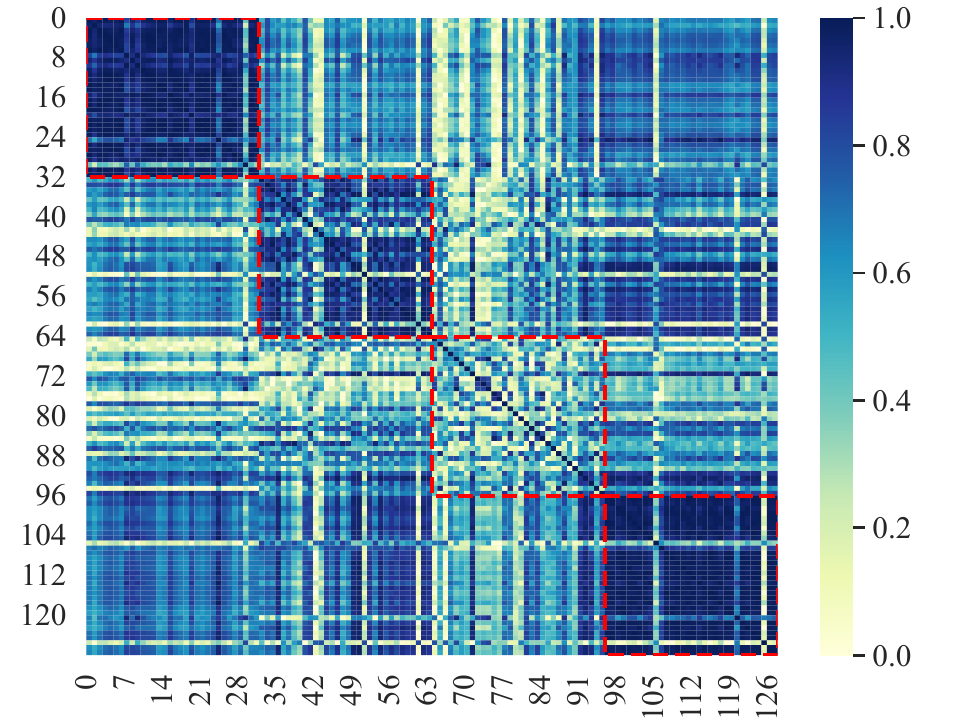}
\caption{DGCL (MUTAG)} \label{fig:7_b}
\end{subfigure}
\begin{subfigure}{0.245\linewidth}
\includegraphics[width=\linewidth]{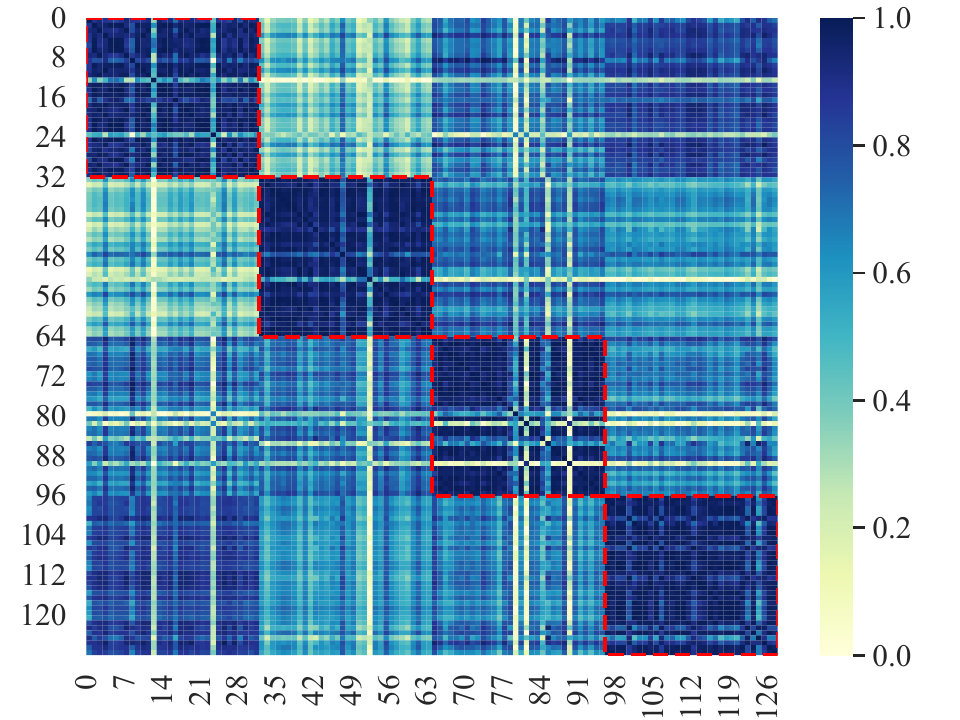}
\caption{\method{} (MUTAG)} \label{fig:7_c}
\end{subfigure}
\begin{subfigure}{0.245\linewidth}
\includegraphics[width=\linewidth]{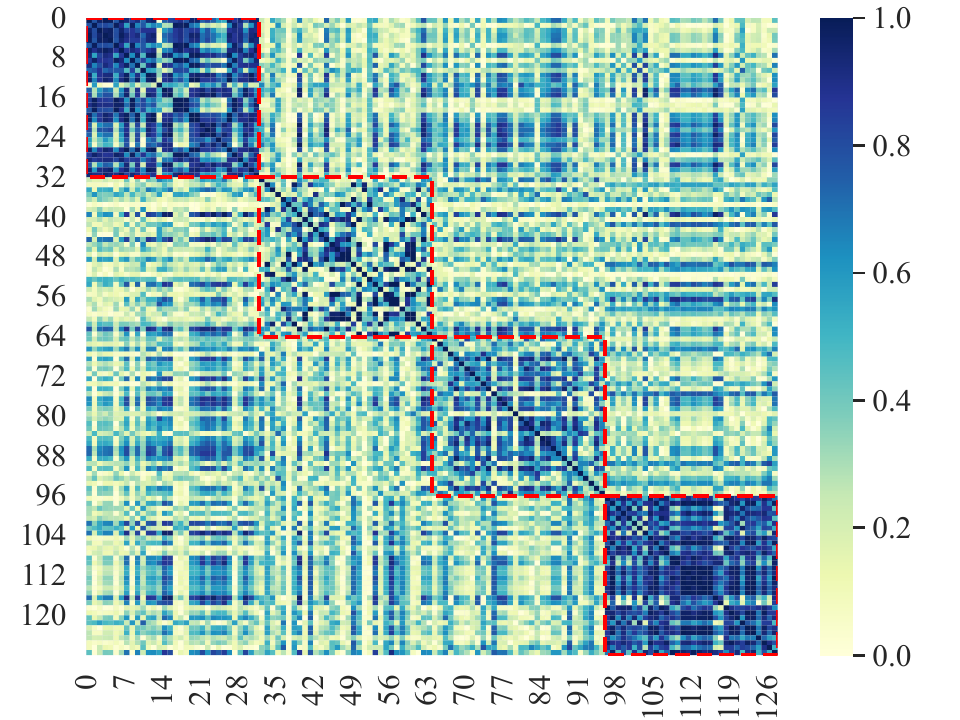}
\caption{\method{} (PROTEINS)} \label{fig:7_d}
\end{subfigure}
% \begin{subfigure}{0.32\linewidth}
% \includegraphics[width=\linewidth]{Figure/fig8_c.pdf}
% \caption{IMDB-BINARY} \label{fig:8_c}
% \end{subfigure}
\caption{Representation correlation analysis, where representations are obtained from the test split in MUTAG and PROTEINS. We use the fine-tuned model of GraphCL and DGCL and the supervised model of our proposed \method{} to get the representation. }
\label{fig:7}
\end{figure*}

\begin{figure*}[t]
    \centering   \includegraphics[width=\linewidth]{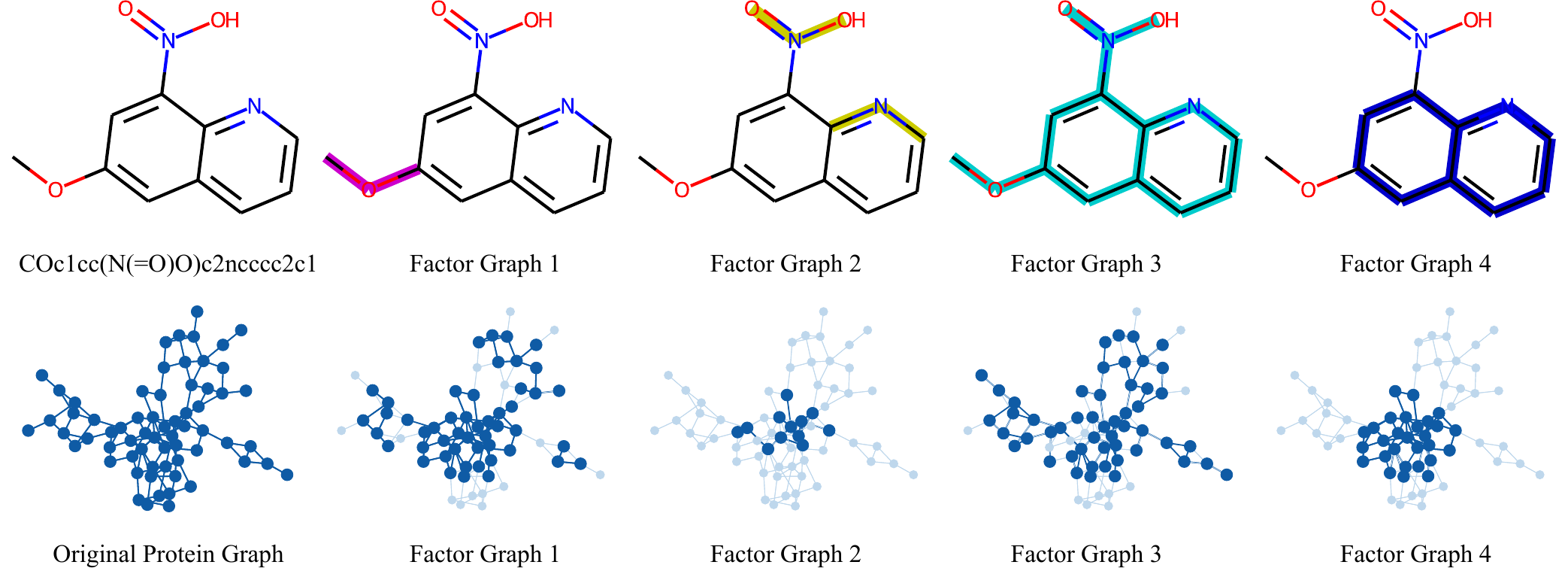}
    \caption{Visualization of disentangled factor graph in MUTAG and PROTEINS. The first column presents the original graph and the second to fifth columns present those factor graphs highlighted with different colors.}
    \label{fig:8} 
\end{figure*}

\subsubsection{Visualization of Representation Correlation}

Besides the quantitative evaluation, we also visualize the absolute correlations between the components of the 128-dimensional graph representations. Fig.~\ref{fig:7} shows the correlations analysis of representation obtained from GraphCL, DGCL and our proposed \method{} with four factor graphs in the MUTAG and PROTEINS dataset. We find that:
\begin{itemize}[leftmargin=*]
\item Compared with the more highly independent representation of DGCL, the learned representation of GraphCL is entangled and the performance is degraded. This indicates that disentangled representation can achieve high performance in the graph classification task. 
\item The representation produced by \method{} demonstrates a distinct block-wise correlation pattern, suggesting that the four factor graphs of \method{} are likely capturing mutually exclusive information. This allows for the extraction of discriminative features for the target task.
\end{itemize}

\subsubsection{Case Study}
To illustrate the disentanglement process more clearly, we present examples of the factor graphs generated by \method{}. Figure~\ref{fig:8} shows the original graph alongside the disentangled factor graphs. The edges of the visualized disentangled factor graphs are highlighted with different colors after we set the coefficient threshold in the original graph. For example, in the MUTAG dataset, the task is to predict the molecule's mutagenicity on Salmonella typhimurium for a collection of nitroaromatic compounds. We can see different parts of a molecule graph playing different roles in prediction. This also justifies the reliability of our generated factor graphs in getting the disentangled graph representation. 

% \subsubsection{Visualization}

\section{Conclusion}
\label{sec:conclusion}
This paper explores the problem of semi-supervised graph classification, a core problem in the field of graph-structured data analysis. For transferring suitable knowledge from the unsupervised model to the supervised model, we propose a novel framework termed \method{}, which learns disentangled representation to capture multiple graph characteristics stemming from different aspects. Specifically, our \method{} consists of a supervised model and an unsupervised model. For both models, we design a disentangled graph encoder to extract factor-wise graph representation and train two models with supervised objective and MI-based constraints, respectively. Then we propose an MI-based disentangled consistency regularization to identify the rationale that aligns well between two models for the current graph classification task and transfer corresponding knowledge semantically. Extensive experiments on ten benchmark graph classification datasets demonstrate the efficacy of our \method{}. However, the number of factor graphs varies among datasets and needs to be searched to find the best value. In future research, we plan to extend our \method{} to more complex semi-supervised classification scenarios and automatically select factor graph numbers in a bi-level optimization framework.
\section*{Acknowledgement}
The paper is supported by ``the Fundamental Research Funds for the Central Universities” in UIBE (Grant No. 23QN02) and the National Natural Science Foundation of China (NSFC Grant Numbers 62306014 and 62276002).
% The authors are grateful to the anonymous reviewers for critically reading this article and for giving important suggestions to improve this article. 

% {\appendix[Proof of the Zonklar Equations]
% Use $\backslash${\tt{appendix}} if you have a single appendix:
% Do not use $\backslash${\tt{section}} anymore after $\backslash${\tt{appendix}}, only $\backslash${\tt{section*}}.
% If you have multiple appendixes use $\backslash${\tt{appendices}} then use $\backslash${\tt{section}} to start each appendix.
% You must declare a $\backslash${\tt{section}} before using any $\backslash${\tt{subsection}} or using $\backslash${\tt{label}} ($\backslash${\tt{appendices}} by itself
%  starts a section numbered zero.)}


\bibliographystyle{IEEEtran}
\bibliography{ref}

\begin{IEEEbiography}
[{\includegraphics[width=1in,height=1.25in]{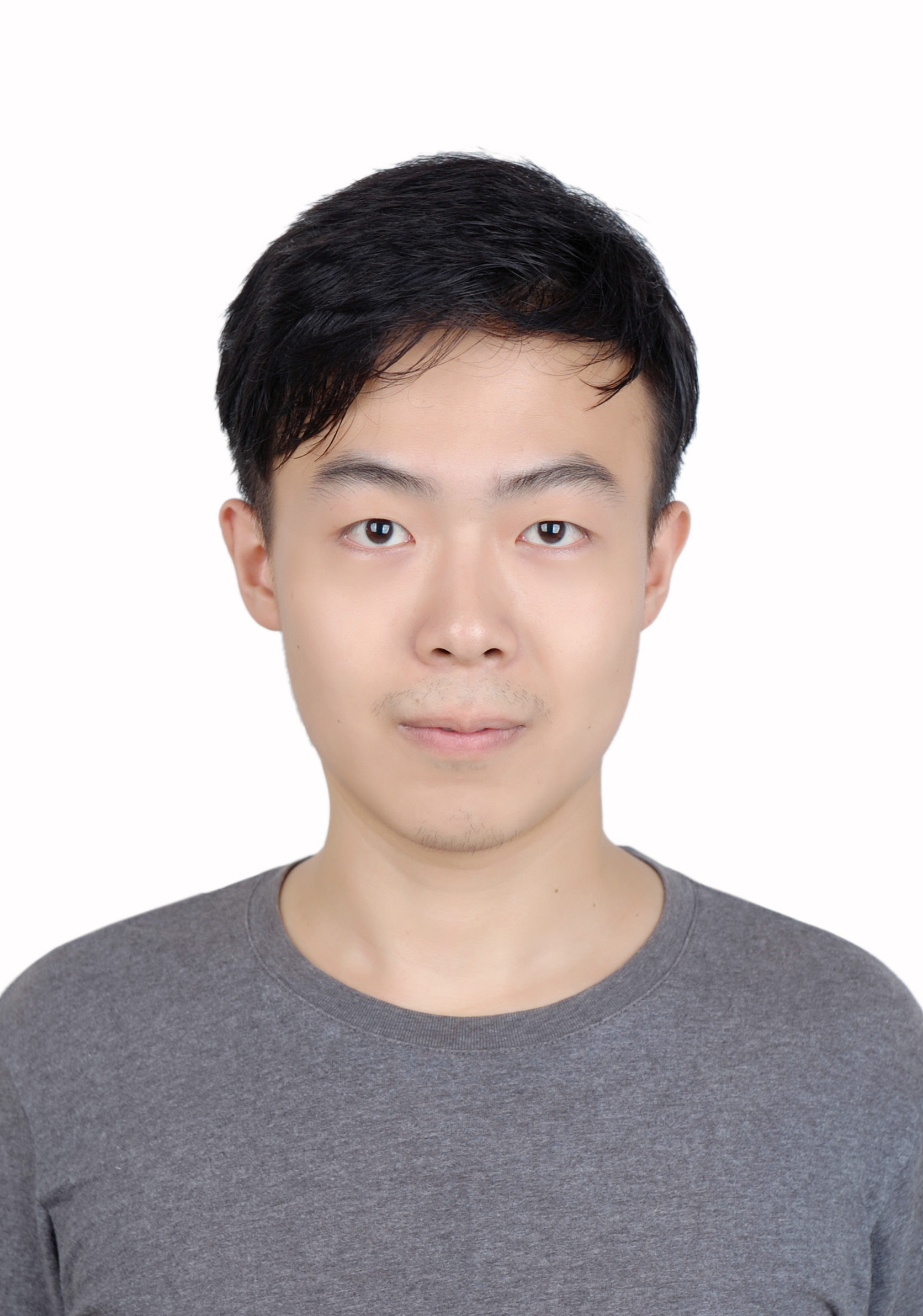}}]
{Yifan Wang} is currently an assistant professor in the School of Information Technology $\&$ Management, University of International Business and Economics. Prior to that, he received his Ph.D. degree in Computer Science from Peking University, Beijing, China, in 2023. He received his M.S. and B.S. degrees in Software Engineering from Northeastern University, Liaoning, China, in 2014 and 2017 respectively. His research interests include graph representation learning, graph neural networks, disentangled representation learning, and corresponding applications such as drug discovery and recommender systems.
\end{IEEEbiography}

\begin{IEEEbiography}
[{\includegraphics[width=1in,height=1.25in]{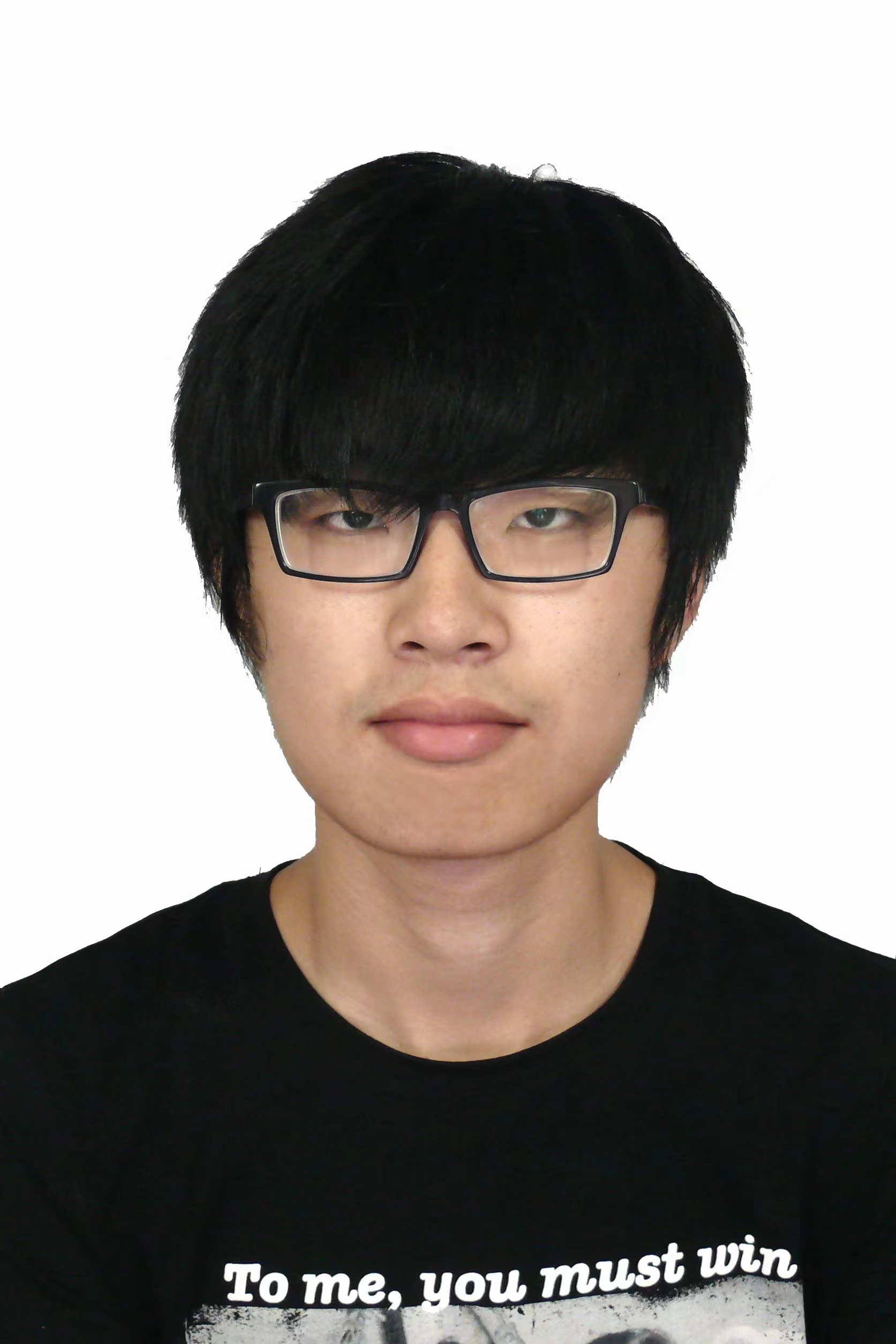}}]
{Xiao Luo} is a postdoctoral researcher in Department of Computer Science, University of California, Los Angeles, USA. Prior to that, he received the Ph.D. degree in School of Mathematical Sciences from Peking University, Beijing, China and the B.S. degree in Mathematics from Nanjing University, Nanjing, China, in 2017. 
His research interests include machine learning on graphs, image retrieval, statistical models and bioinformatics. 
\end{IEEEbiography}

% \begin{IEEEbiography}
% [{\includegraphics[width=1in,height=1.25in]{bio/Liu.jpg}}]
% {Luchen Liu} is currently a post-doctoral research fellow in Computer Science at Peking University. He received the Ph.D. degree in Computer Science from Peking University in 2020. His current research interests lie primarily in the area of deep learning for temporal graph data and interdisciplinary applications such as intelligent healthcare and quantitative investment.
% \end{IEEEbiography}

\begin{IEEEbiography}
[{\includegraphics[width=1in,height=1.25in]{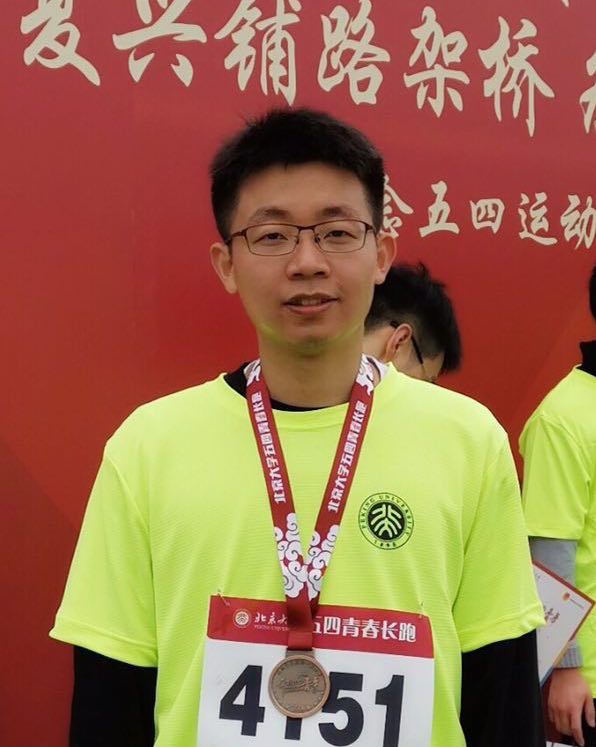}}]
{Chong Chen} is currently a research scientist in Terminus Group. He received the B.S. degree in Mathematics from Peking University in 2013 and the Ph.D. degree in Statistics from Peking University in 2019 under the supervision of Prof. Ruibin Xi. His research interests include image understanding, self-supervised learning, and data mining.
\end{IEEEbiography}

\begin{IEEEbiography}
[{\includegraphics[width=1in,height=1.25in]{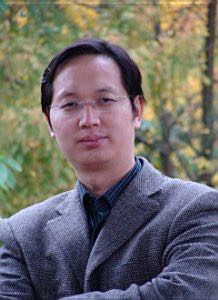}}]
{Xian-Sheng Hua} (Fellow, IEEE) received the B.S. and Ph.D. degrees in applied mathematics from Peking University, Beijing, in 1996 and 2001, respectively. In 2001, he joined Microsoft Research Asia, Beijing, as a Researcher, and has been a Senior Researcher at Microsoft Research Redmond, Redmond, WA, USA, since 2013. He became a Researcher and the Senior Director of Alibaba Group, Hangzhou, China, in 2015. He has authored or coauthored over 250 research articles and has filed over 90 patents. His research interests include multimedia search, advertising, understanding, and mining, pattern recognition, and machine learning. He was honored as one of the recipients of MIT Technology Review Innovators Under 35 Asia Pacific (MIT35). He served as a Program Co-Chair for the ACM Multimedia 2012, the IEEE International Conference on Multimedia and Expo (ICME) 2012, IEEE ICME 2013, and on the Technical Directions Board for the IEEE Signal Processing Society. He is an ACM Distinguished Scientist.
\end{IEEEbiography}

\begin{IEEEbiography}
[{\includegraphics[width=1in,height=1.25in]{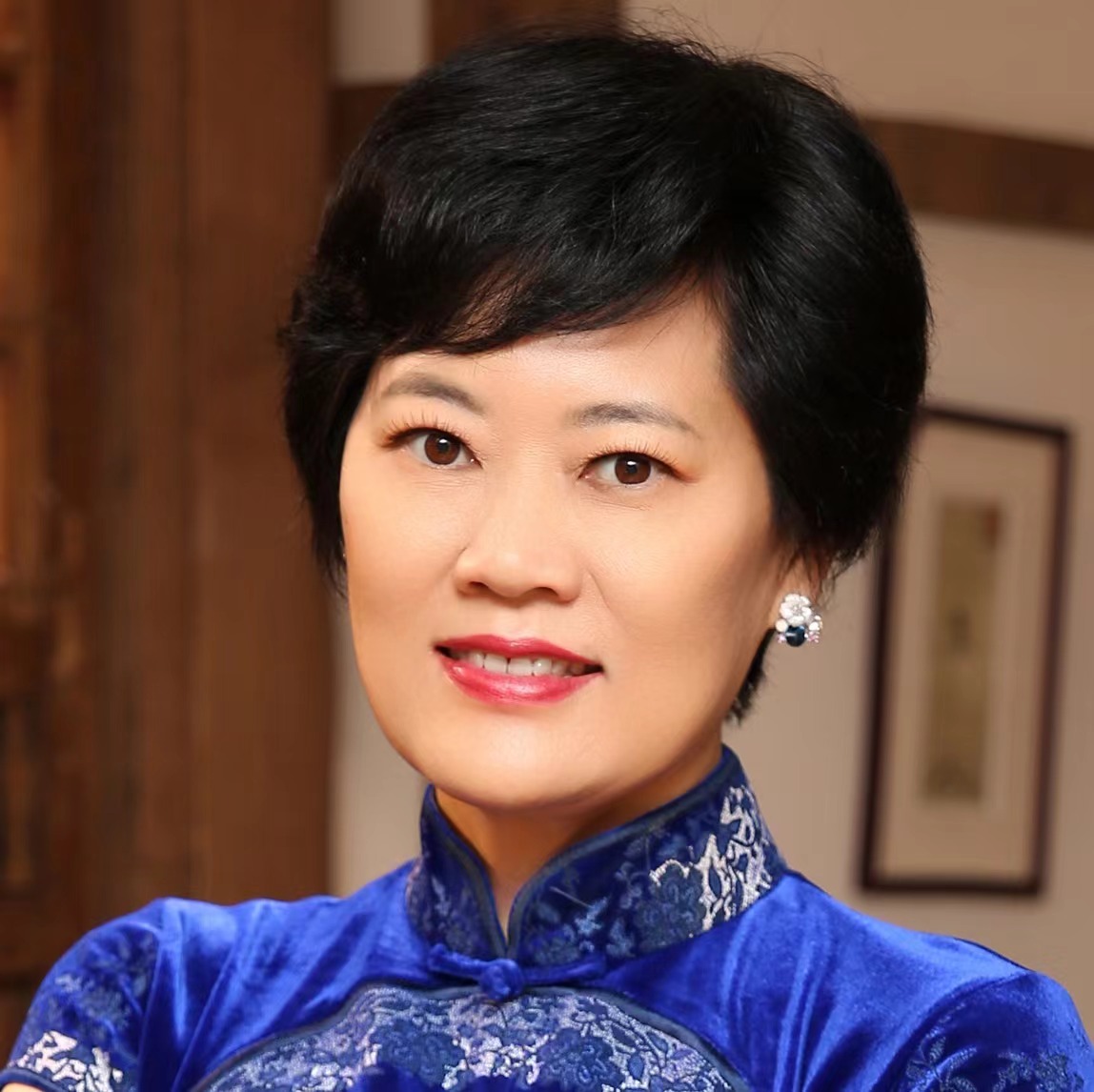}}]
{Ming Zhang} received her B.S., M.S. and Ph.D. degrees in Computer Science from Peking University respectively. She is a full professor at the School of Computer Science, Peking University. Prof. Zhang is a member of Advisory Committee of Ministry of Education in China and the Chair of ACM SIGCSE China. She is one of the fifteen members of ACM/IEEE CC2020 Steering Committee. She has published more than 200 research papers on Text Mining and Machine Learning in the top journals and conferences. She won the best paper of ICML 2014 and best paper nominee of WWW 2016. Prof. Zhang is the leading author of several textbooks on Data Structures and Algorithms in Chinese, and the corresponding course is awarded as the National Elaborate Course, National Boutique Resource Sharing Course, National Fine-designed Online Course, National First-Class Undergraduate Course by MOE China.
\end{IEEEbiography}

\begin{IEEEbiography}
[{\includegraphics[width=1in,height=1.25in]{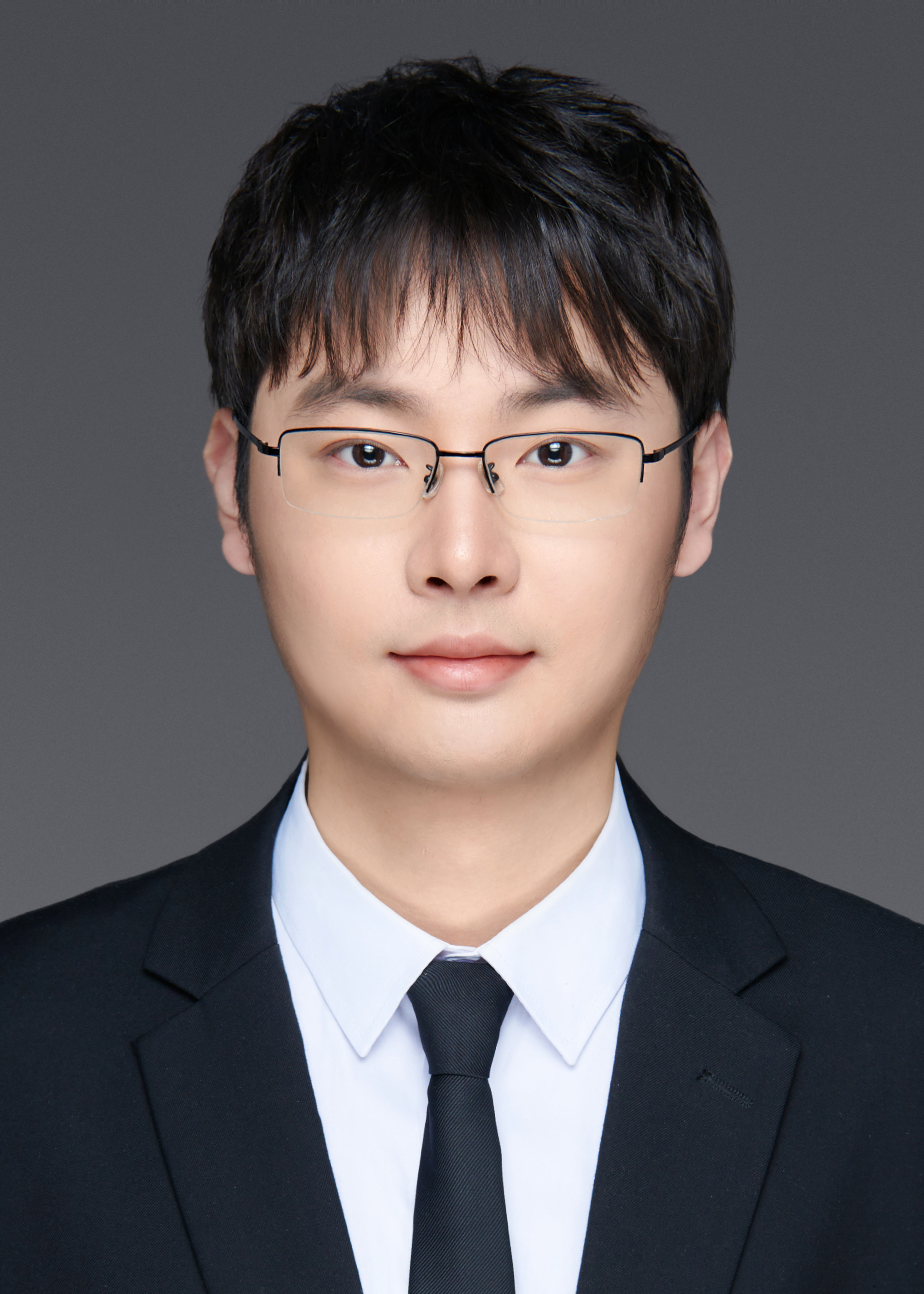}}]
{Wei Ju} is currently an associate professor with the College of Computer Science, Sichuan University, Chengdu, China. Prior to that, he worked as a postdoc research fellow and received his Ph.D. degree in the School of Computer Science from Peking University, Beijing, China, in 2022. He received the B.S. degree in Mathematics from Sichuan University, Sichuan, China, in 2017. His current research interests lie primarily in the area of machine learning on graphs including graph representation learning and graph neural networks, and interdisciplinary applications such as recommender systems, bioinformatics, drug discovery and knowledge graphs. He has published more than 40 papers in top-tier venues and has won the best paper finalist in IEEE ICDM 2022.
\end{IEEEbiography}
% \newpage

 

% \end{IEEEbiography}



% \vfill

\end{document}